\documentclass{article}
\usepackage{xcolor}

\usepackage{times}
\usepackage[T1]{fontenc}    %
\usepackage{url}            %
\usepackage{booktabs}       %
\usepackage{nicefrac}       %
\usepackage{microtype}      %
\usepackage{graphics,color}

\usepackage{algorithm}
\usepackage{algorithmic}
\usepackage{enumitem}

\usepackage{amsfonts}       %
\usepackage{amsmath}       %
\usepackage{amssymb}

\newcommand{\Fig}[1]{Figure~\ref{#1}}  %
\newcommand{\fig}[1]{Fig.~\ref{#1}}    %

\newcommand{\tab}[1]{Table~\ref{#1}}

\newcommand{\eqn}[1]{Eq.~\ref{#1}} %
\renewcommand{\sec}[1]{Sec.~\ref{#1}} %
\newcommand{\supp}[1]{Suppl.~\ref{#1}}

\usepackage{xspace}
\makeatletter
\DeclareRobustCommand\onedot{\futurelet\@let@token\@onedot}
\def\@onedot{\ifx\@let@token.\else.\null\fi\xspace}

\makeatother

\definecolor{ourblue}{rgb}{0.368,0.507,0.71}
\definecolor{ourorange}{rgb}{0.881,0.611,0.142}
\definecolor{ourgreen}{rgb}{0.56,0.692,0.195}
\definecolor{ourred}{rgb}{0.923,0.386,0.209}
\definecolor{ourviolet}{rgb}{0.528,0.471,0.701}
\definecolor{ourbrown}{rgb}{0.772,0.432,0.102}
\definecolor{ourlightblue}{rgb}{0.364,0.619,0.782}
\definecolor{ourdarkgreen}{rgb}{0.572,0.586,0.}

\definecolor{ourcyan2}{rgb}{0.125,0.722,0.804}
\definecolor{ourred2}{rgb}{0.863,0.184,0.047}
\definecolor{ouryellow2}{cmyk}{0,0.16,1.0,0.07}
\definecolor{ourviolet2}{cmyk}{0.55,0.56,0,0.47}
\definecolor{ourorange2}{cmyk}{0,0.46,0.89,0.11}

\usepackage{amsthm}
\usepackage{graphicx}
\usepackage{mathtools}
\usepackage{caption}
\usepackage{subcaption}
\usepackage{wrapfig}
\usepackage{xspace}
\usepackage{enumitem}
\usepackage{xr-hyper}
\makeatletter
\newcommand*{\addFileDependency}[1]{%
  \typeout{(#1)}
  \@addtofilelist{#1}
  \IfFileExists{#1}{}{\typeout{No file #1.}}
}
\makeatother

\usepackage[skip=4pt]{parskip}

\newcommand{\thetades}{\theta_\mathrm{i}^\mathrm{des}}
\newcommand{\thetadeship}{\theta_\text{hp}^\mathrm{des}}
\newcommand{\thetadesknee}{\theta_\text{kn}^\mathrm{des}}
\newcommand{\thetadesankle}{\theta_\text{an}^\mathrm{des}}
\newcommand{\thetadotdes}{\dot{\theta}_\mathrm{i}^\mathrm{des}}
\newcommand{\fce}{f_\text{CE}}

\usepackage{listings}

\newcommand*{\stim}{u}

\newcommand*{\act}{a}

\newcommand*{\lmtui}{l_{\text{MTU},i}}
\newcommand*{\dlmtui}{\dot{l}_{\text{MTU},i}}
\newcommand*{\lmtu}{l_\text{MTU}}
\newcommand*{\dlmtu}{\dot{l}_\text{MTU}}
\newcommand*{\lcei}{l_{\text{CE},i}}
\newcommand*{\lce}{l_\text{CE}}
\newcommand*{\dlcei}{\dot{l}_{\text{CE},i}}
\newcommand*{\dlce}{\dot{l}_\text{CE}}

\newcommand*{\fmtui}{f_{\text{MTU},i}}
\newcommand*{\fmtu}{f_\text{MTU}}

\usepackage{amsmath}
\usepackage{pgfplots}
\usepackage{varwidth}
\usepackage{tikz}
\usetikzlibrary{external,shapes,arrows,fit,positioning}
\newlength{\figW}
\newlength{\figH}
\usepackage{pdfpages}
\usepackage{caption,subcaption}
\usetikzlibrary{positioning}
\usepackage{xcolor}
\usepackage{amssymb}

\usepackage[percent]{overpic}

\usepackage{graphicx}

\definecolor{orange}{HTML}{f28e2b}
\definecolor{blue}{HTML}{4e79a7}
\definecolor{red}{HTML}{d62728} %
\definecolor{purple}{HTML}{b07aa1}
\definecolor{pink}{HTML}{E28BF3}

\definecolor{green}{HTML}{59a14f}

\newcommand{\biped}{Biped\xspace}
\newcommand{\armdemoa}{ArmDemoa\xspace}
\newcommand{\allmin}{FullBody\xspace}
\newcommand{\arm}{Arm\xspace}
\newcommand{\armmujoco}{ArmMuJoCo\xspace}
\usepackage[final]{corl_2022} %

\title{Learning with Muscles: Benefits for Data-Efficiency and Robustness in Anthropomorphic Tasks}

\author{%
Isabell Wochner\thanks{Equal contribution. $\dagger$ Equal contribution.} ~$^{1}$
  \And
  Pierre Schumacher$^{*2,3}$
  \And
  Georg Martius$^{2}$
  \AND
  Dieter Büchler$^{2}$
  \And
  Syn Schmitt$^{\dagger 1}$
  \And
  Daniel F.B. Haeufle$^{\dagger 1,3}$ 
   \AND
   \normalfont{$^{1}$Institute for Modelling and Simulation of Biomechanical Systems, University of Stuttgart, Germany} \\
   $^{2}$Max Planck Institute for Intelligent Systems, Tübingen, Germany\\
   $^{3}$Hertie-Institute for Clinical Brain Research, University of Tübingen, Germany}
   
\begin{document}
\maketitle
\begin{abstract}
	Humans are able to outperform robots in terms of robustness, versatility, and learning of new tasks in a wide variety of movements. 
	We hypothesize that highly nonlinear muscle dynamics play a large role in providing inherent stability, which is favorable to learning. 
	While recent advances have been made in applying modern learning techniques to muscle-actuated systems both in simulation as well as in robotics, so far, no detailed analysis has been performed to show the benefits of muscles when learning from scratch.
	Our study closes this gap and showcases the potential of muscle actuators for core robotics challenges in terms of data-efficiency, hyperparameter sensitivity, and robustness \footnote{See \url{https://sites.google.com/view/learning-with-muscles} for code and videos.}.
\end{abstract}
\begin{figure}[h]
	\centering
 	\includegraphics[height=3.75cm]{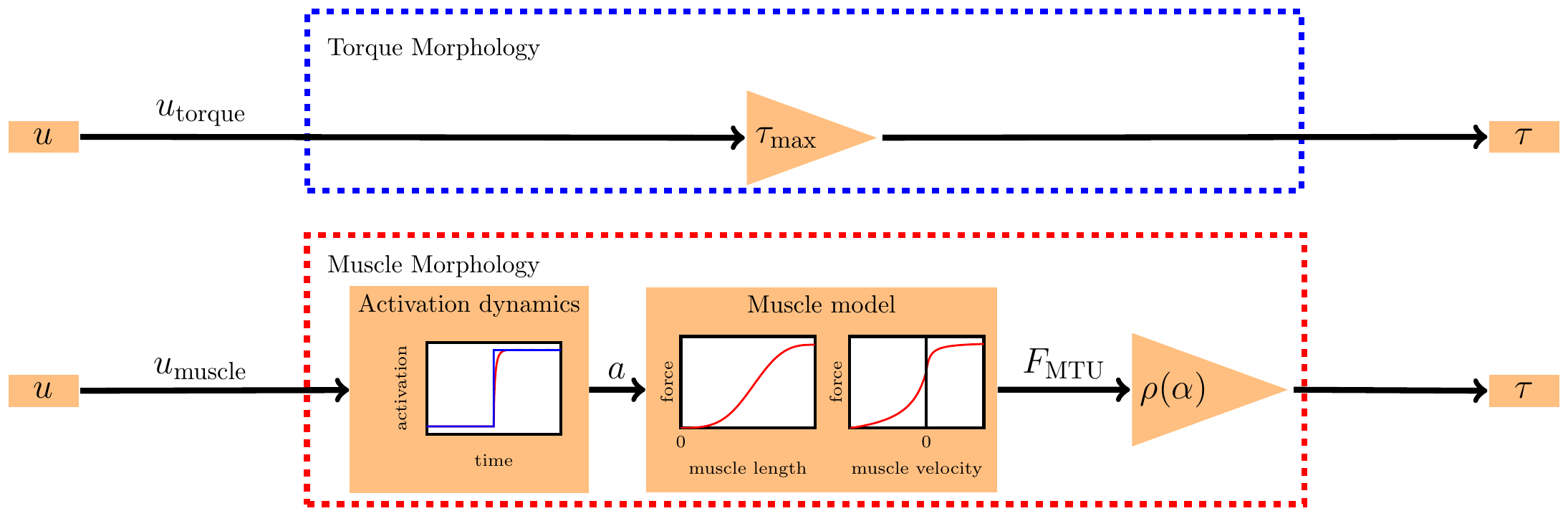}
 	\caption{Key differences between torque actuator morphology and muscle actuator morphology.} \label{fig:muscle_torq_morph}
 	\label{fig:muscle_flva_plot}
\end{figure}
\section{Introduction}
Recent developments in new learning methods allow the generation of complex anthropomorphic motions such as grasping, jumping or hopping in robotics. However, current systems still struggle with real-world scenarios beyond the narrow conditions of laboratory experiments. Humans, on the other hand, are capable of quickly adapting to uncertain, complex, and changing environments in a sheer endless variety of tasks. One key difference between biological and robotic systems lies in their actuator morphology: robotic drives are mostly designed to yield a linear relation between control signal and output torque. In contrast, muscles have complex nonlinear characteristics. 

It has already been demonstrated, that muscular nonlinearities may provide a benefit for stability and robustness, especially under environmental uncertainties or perturbations~\citep{wagner1999stabilizing,eriten2009rigorous,Brandle2019a}. A benefit over linear torque actuator morphology has been observed in computer simulations by exchanging the actuator morphology (similar to Fig. \ref{fig:muscle_flva_plot}) in otherwise identical anthropomorphic tasks like reaching \citep{Stollenmaier2020} or locomotion \citep{van1993contribution,Haeufle2010a,gerritsen1998intrinsic,john2013stabilisation}. Similarly, reduced demand on the information processing capacity has been shown for muscles when compared to torque actuator morphology~\citep{full1999templates,holmes2006dynamics,blickhan2007intelligence,haeufle2014quantifying,haeufle2020muscles}. This opens the question whether muscular morphology could also be beneficial for robustness and data-efficiency in the process of \textit{learning} movement control.

Recent advances in deep learning facilitated the generation of complex movements like point-reaching ~\cite{fischerReinforcementLearningControl2021a, joosReinforcementLearningMusculoskeletal2020a,myosuite2022, schumacherdeprl2022} and locomotion~\cite{barbera2021ostrichrl, kidzinski_learning_2018-2,kidzinski_artificial_2019-1, Song2021,schumacherdeprl2022} in simulations with muscular actuator morphology. %
In the real world, optimization and learning approaches can also find controllers for robotic systems with pneumatic muscles exhibiting somewhat muscle-like actuator morphology \cite{driess2018learning, buchler2019learning}. These examples demonstrate that learning and optimization methods \textit{can} control muscle-driven systems and may enable benefits such as safe learning and robustness \cite{buchler2019learning}. However, investigating advantages of nonlinear muscular actuator morphology over linear torque actuator morphology requires a direct comparison of both, which is---to our knowledge---missing in the literature.

While Peng et al.~\cite{peng2017learning} performed a comparative analysis of different actuator morphologies, their work was focused on replicating reference trajectories. 
In contrast, we learn behaviors without demonstrations, provide extensive hyperparameter ablations and not only employ RL, but also other optimization methods applied to complex 3D models.

The purpose of this study is to test if learning with muscular actuator morphology is more data-efficient and results in more robust performance as compared to torque actuator morphology when learning from scratch. We investigate this in a very broad approach: we employ different learning strategies on multiple anthropomorphic models for multiple variants of reaching and locomotion tasks solved in physics simulators of differing levels of detail. This provides new and comprehensive evidence of the beneficial contribution of muscular morphology to the learning of diverse movements.

\section{Morphological difference between torque and muscle actuators}
\label{sec:muscles}
In contrast to idealized torque actuators, where torque is simply proportional to the control signal $u_{\text{torque}} \in [-1,1]$,
\begin{equation}
	\tau = \tau_{\text{max}}\,u_{\text{torque}} 
\end{equation}
muscular force output nonlinearly depends on the muscle control signal $u_{\text{muscle}}$, the muscle length $l_\text{MTU}$ and contraction velocity $\dot{l}_\text{MTU}$. These biologically observed dependencies can be predicted by so-called \emph{Hill-type} muscle models \citep{Siebert2014}. 
In a nutshell, the model captures biochemical processes transforming muscle stimulation $u_{\text{muscle}} \in [0,1]$ to the force-generating calcium ion activity $a$. This can be modeled by a first-order differential equation of the form \citep{Rockenfeller2015}
\begin{equation}
	\dot{a} = f_a(u_{\text{muscle}} - a)
\end{equation}
which induces low-pass filter characteristics (Fig.~\ref{fig:muscle_flva_plot}).
The model further captures the nonlinear \textit{force-length} and \textit{force-velocity relations} \citep{Siebert2014}. The \textit{force-length relation} is characterized by a positive slope (increasing force with increasing muscle fiber length) in the typical operating range of biological muscle fibres (Fig.~\ref{fig:muscle_flva_plot}). The \textit{force-velocity} relation is characterized by decreasing force for faster shortening velocities and increasing force if the muscle is externally stretched against its contraction direction (Fig.~\ref{fig:muscle_flva_plot}). A lever arm $\rho(\alpha)$ translates joint angle $\alpha$ into muscle-tendon-unit length $l_{\text{MTU}}$ and muscle force into joint torque
\begin{equation}
	\tau = \sum_{i=1}^{N} \rho_{i}(\alpha) f_{\tau}\left(l_{\text{MTU},i}(\alpha), \dot{l}_{\text{MTU},i}(\dot{\alpha}), a_{i}\right). 
\end{equation}
for $N$ muscles which span a joint---typically at least two in an antagonistic arrangement.

In practice, we employ two different muscle models: A detailed one with more physiological details, contained in demoa~\citep{darus-2550_2022}, and a simpler model that efficiently adds muscular properties to existing MuJoCo~\cite{todorovMuJoCoPhysicsEngine2012a} simulations without sacrificing computational speed. See \supp{supp:muscles} for details.
\section{Methods}
\begin{table}
	\centering \small
	\caption{\textbf{Overview of all models and tasks}}
	{\begin{tabular}{llccc} \toprule
			Model & Task & Control & Environment  \\  \midrule
			\armmujoco & precise reaching & RL & MuJoCo  \\   
			\armmujoco & fast reaching & RL & MuJoCo \\   
			\armdemoa  & smooth point-reaching & opt. control, MPC & demoa \\
			\armdemoa  & hitting ball with high-velocity & opt. control, MPC & demoa \\
			\biped & hopping & RL & MuJoCo \\
			\allmin  & squatting & opt. control, MPC & demoa \\
			\allmin  & high-jumping & opt. control, MPC & demoa\\ \bottomrule
	\end{tabular}}
	\label{tab:overviewmodelstasks}
\end{table}
\begin{figure}
	\centering
	\begin{subfigure}[b]{0.24\textwidth}
		\centering
		\includegraphics[width=0.55\textwidth]{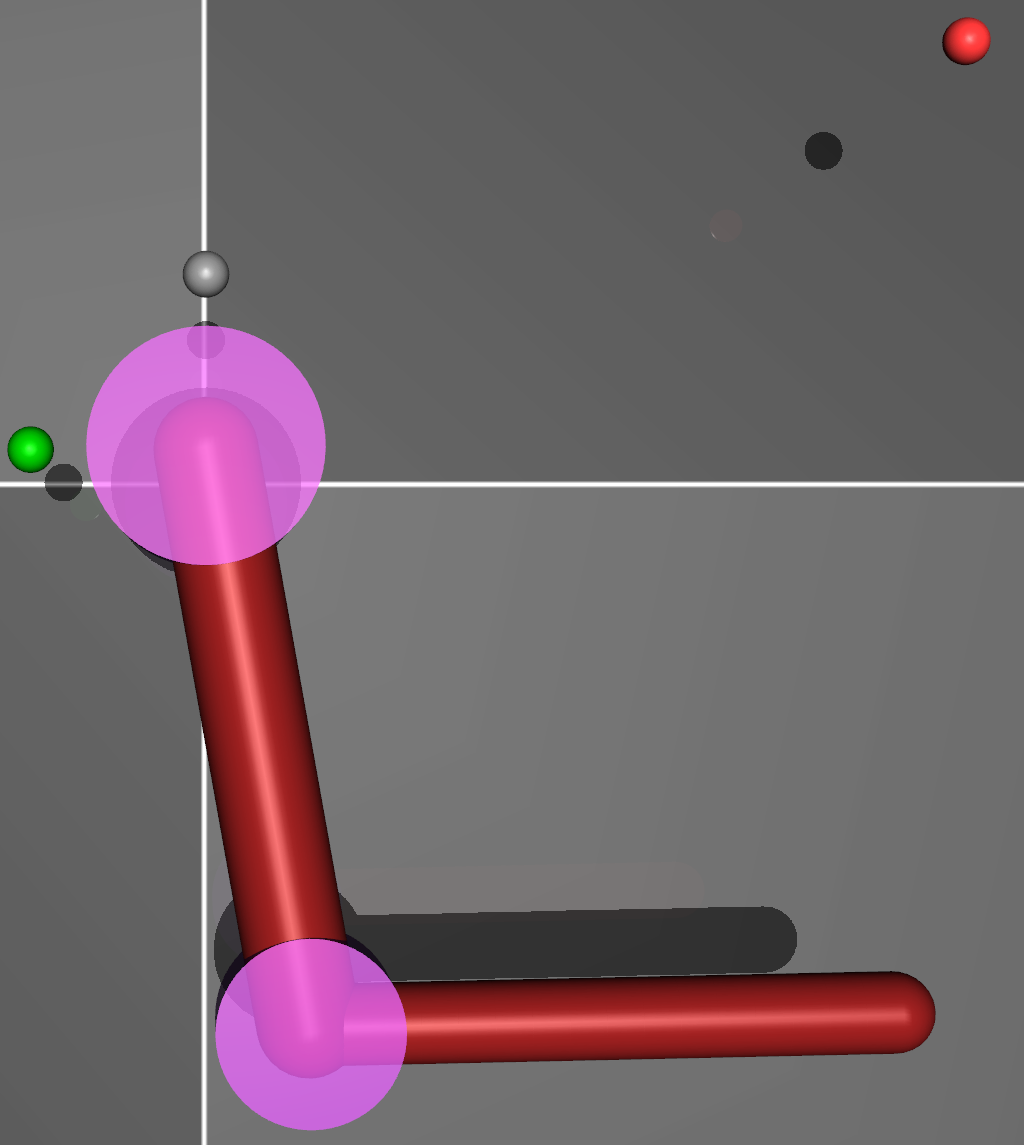}
		\caption{\armmujoco}
		\label{subfig:armdemoa}
	\end{subfigure}
	\hfill
		\begin{subfigure}[b]{0.24\textwidth}
		\centering
		\includegraphics[width=0.6\textwidth]{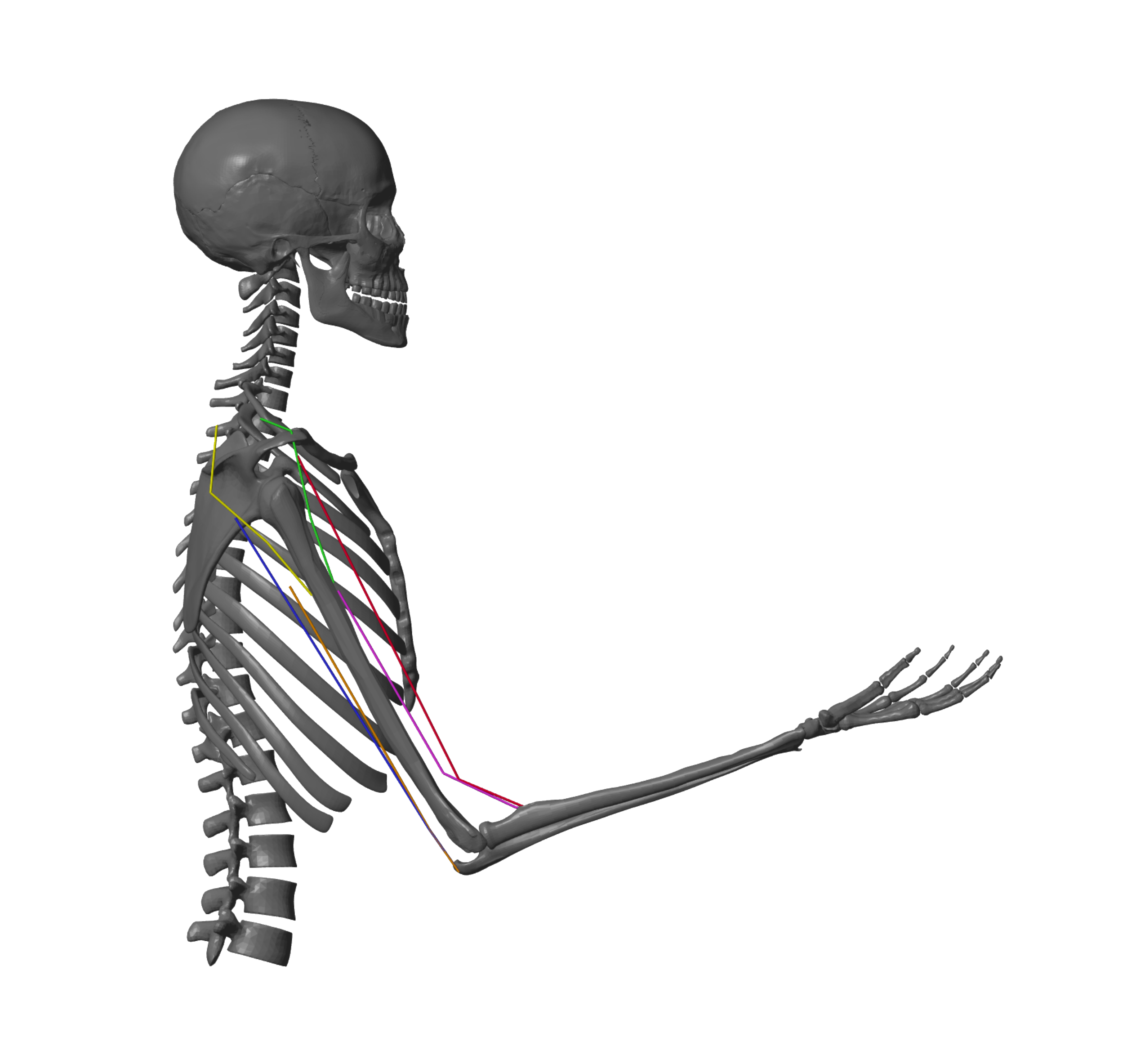}
		\caption{\armdemoa}
		\label{subfig:armmujoco}
	\end{subfigure}
	\hfill
	\begin{subfigure}[b]{0.24\textwidth}
		\centering
		\includegraphics[width=0.62\textwidth]{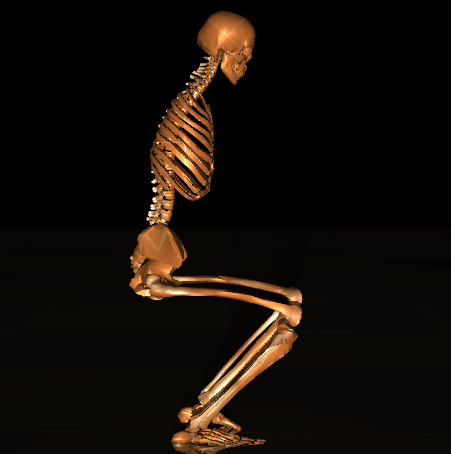}
		\caption{\biped}
		\label{subfig:biped}
	\end{subfigure}
	\hfill
	\begin{subfigure}[b]{0.24\textwidth}
		\centering
		\includegraphics[width=0.2\textwidth]{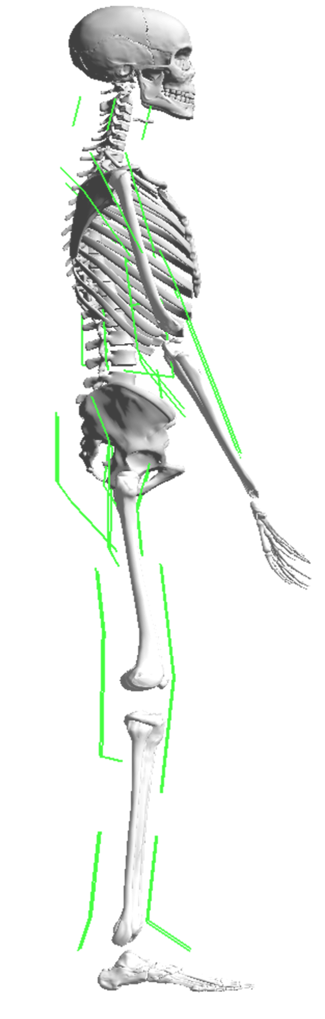}
		\caption{\allmin}
		\label{subfig:allmin}
	\end{subfigure}
	\caption{Models used for the experiments.}
	\label{fig:models}
\end{figure}
\subsection{Learning approaches for movement control}
We test if muscle actuator morphology facilitates learning by applying state-of-the-art learning algorithms covering an extensive range of approaches currently used in robotics. The common thread of the selected algorithms lies in their independence of the actuator morphology: this allows us to easily exchange idealized torque actuator morphology with muscle actuator morphology. We choose optimal control, model-predictive control and reinforcement learning as learning approaches.
\paragraph{Optimal control (OC)}
The control problem with horizon $N$ can be defined as:
\begin{align}
	\min_{\pi_{k}} J = \min_{\pi_k} \sum_{k=0}^{N} l(x(k),u(k),k),\qquad	\text{subject to}\ \  x(k+1) &= f(x(k),u(k),k),& \nonumber\\
	u(k) &= \pi_k(x(0),...,x(k)).& \label{eq:policy}
\end{align}
where $x(k)\in \mathbb{R}^{n_x}$ denotes the current state at time $k$, and $u(k)\in \mathbb{R}^{n_u}$ is the applied input at time $k$. Furthermore, $l$ specifies the cost function, and $f$ denotes the system dynamics. To optimize for the best control policy, we use the covariance matrix adaptation evolution strategy (CMA-ES) \citep{hansen2003reducing} in the optimal control case (open-loop strategy). CMA-ES is a derivative-free algorithm and widely used in machine learning. It combines different learning mechanisms for adapting the parameters of a multivariate normal distribution. Note, that we choose the same optimization parameters for both actuator morphologies to allow for a fair comparison even though the number of decision variables $n_u$ is always larger in the muscle-actuated case due to the antagonistic setup. 
\paragraph{Model predictive control (MPC)}
 In MPC, we solve the control problem in a receding-horizon fashion with varying prediction horizons and recursively apply only the first element of the predicted optimal control sequence $u(0)$ (closed-loop strategy). We employ a warm start procedure using the CMA-ES optimizer and afterwards start the MPC routine with the local optimizer BOBYQA \citep{powell2009bobyqa}. The parameters of the optimizers are either varied (see \sec{sec:results}) or given in \supp{supp:implementation}.
\paragraph{Reinforcement learning (RL)}
RL allows learning of reusable feedback controllers. Instead of \textit{minimizing} a cost function (see \eqn{eq:policy}), conventionally the discounted expected reward is \textit{maximized}: 
\begin{equation}
    \max_{\pi} J =\max_{\pi}\,\mathbb{E}\left[\sum_{k=0}^{N-1}\gamma^{k-1}\,r(k)\right]
\end{equation} 
where $r(k)$ is the reward at time $k$, $\pi$ is a control policy and $\gamma\in[0,1]$ is a discount factor such that long-term rewards are weighted less strongly. RL consequently solves a similar problem to MPC, but the resulting controllers are able to act in closed-loop fashion without being given an explicit prediction model. For the point-reaching tasks, we additionally employ goals $g$ characterizing the desired hand position, which then constitutes an additional dependence of the reward function. The aim of the learning process is to learn a controller policy $\pi(u(k)|x(k))$ that takes as input the current sensor values, or state x(k), and outputs a control signal, or action, $u(k)$ such that a task is solved. In practice, we use the RL algorithm MPO~\cite{abdolmaleki2018maximum}, implemented in TonicRL~\cite{pardo2020tonic}.
\subsection{Models}
\paragraph{\arm{}}
\label{sec:arm26}
The \arm{} model (\fig{fig:models} a, b) consists of two segments connected with hinge joints (2 joints total) moving against gravity. The \armmujoco\,\cite{todorovMuJoCoPhysicsEngine2012a,schumacherdeprl2022} model contains four muscles, two for each joint. In the muscle-actuated case in \armdemoa~\cite{darus-2871_2022,wochner2020optimality}, six Hill-Type muscles generate forces: two muscles for the shoulder and two for the elbow joint, plus two biarticular muscles acting on both joints. All joints are individually controllable.
\vspace{-5pt}
\paragraph{\biped}
\label{sec:biped}
We converted the geometrical model of an OpenSim bipedal human without arms~\cite{kidzinski_learning_2018-2} for use in MuJoCo. The model, consisting of 7 controllable joints (lower back, hips, knees, ankles) moves in a 2D plane. Each joint is actuated by two antagonistic muscles or one torque actuator.
\vspace{-5pt}
\paragraph{\allmin}
\label{sec:allmin}
The \allmin model \citep{walter2021geometry,darus-2982_2022} consists of two legs and an upper body including arms based on a human skeletal geometry. It consists of 8 controllable joints (ankles, knees, hips, lumbar and cervical spine) in 3D, and 14 movable joints in total including the arms. Each controllable joint was either actuated by two antagonistic muscles (muscle-actuated case) or by one idealized torque actuator (torque-actuated case). 
For more details, we refer to \supp{supp:models}.\\
All models and their respective physics differential equations were solved with variable time step (max. time step 0.001s) in demoa and fixed time step (0.005s) in MuJoCo.
\subsection{Objectives and rewards}
We choose anthropomorphic movement objectives which are highly relevant for robotic applications. We expect that muscular actuator morphology provides benefits for such tasks. All task formulations allow application in muscle and torque actuator morphologies with an identical reward or objective function. For a precise formulation of the used functions and conditions, see \supp{supp:tasks}.
\label{sec:costfuncs}
\vspace{-5pt}
\paragraph{Smooth point-reaching}
This task encourages \textit{smooth} point-reaching. Therefore, the objective minimizes the L2-error between the desired and current joint angle while penalizing the angle velocity and jerk to ensure a smooth motion. The desired angle is $90^{\circ}$ for both the shoulder and the elbow joint, as this requires a large motion.
\vspace{-5pt}
\paragraph{Precise point-reaching}
Similar to~\cite{haeufle2020muscles}, we incentivize reaching a random hand position in a pre-determined rectangle, while minimizing the distance of the end effector to the goal. We specifically add a reward term that gives a much larger reward for precise motions that reach the center of the target area. The episode does not terminate until a time limit of 1000 steps elapses.
\vspace{-5pt}
\paragraph{Fast point-reaching}
The same objective as for precise point-reaching is used, but the episode terminates if the target is reached, incentivizing reaching speed over precision.
\vspace{-5pt}
\paragraph{High-velocity ball serve}
A ball is dropped in front of the \arm{} model and the controller learns to hit the ball to achieve maximum ball velocity.
\vspace{-5pt}
\paragraph{Squatting}
This squatting objective is taken from \cite{walter2021geometry} and encourages desired hip, knee, and ankle angles for a squatting position.
\vspace{-5pt}
 \paragraph{Maximum height jump}
 The objective for the high-jumping task is taken from \citep{pandy1990optimal} and maximizes the position and velocity of the centre of mass of the human body model at the time of lift-off. The model is initialized to start from a squatting position.
\vspace{-5pt}
\paragraph{Hopping}
We developed an objective based on the z-axis velocity of the center of mass~(COM) of the system that encourages periodic hopping with maximum height. The episode terminates if extreme joint angles are exceeded.
\section{Results}
\label{sec:results}
In the following, we present three major results for the investigated approaches and environments: (\textbf{1}) Muscle-like actuators in general improve data-efficiency compared to torque-actuators. (\textbf{2}) The investigated learning and optimization algorithms exhibit greater robustness to hyperparameter variations when applied to muscle-driven systems. (\textbf{3}) The motions and controllers obtained from the muscular morphology are more robust against force perturbations that were not present during learning. We average results over $5$ and $8$ random seeds for OC/MPC and RL respectively.
\subsection{Data efficiency: Learning with limited resources}
Robotics applications in real-world scenarios often suffer from limited resources, which holds true for training and inference time. Therefore, we investigate the advantages of muscle-like actuator morphology in terms of overall learning efficiency and temporal control resolution.
\paragraph{Advantages of muscular morphology} Smooth and precise point-reaching generally require more data with torque-driven systems, as seen in \fig{fig:data_efficiency}. The performance of the muscle actuator, in contrast to torque morphology, varies very little for different settings of the temporal control resolution $c$. Precise reaching with RL also results in stable performance with fewer training iterations, and a very small standard deviation across runs. Similar findings are seen for the squatting and hopping task, where muscle-actuators achieve better data-efficiency and smaller variation across runs and are able to find a good-enough optimum with fewer iterations.
\paragraph{Advantages of torque morphology} 
In tasks requiring fast and strong motions, without emphasis on stabilization, we find torque actuators to hold certain advantages. In ball hitting and fast reaching, the torque cases show similar or smaller variance, even though both actuators perform well for singular runs. The high-jumping task, where only a strong, swift motion is required to launch the system upwards, is solved much faster in the torque case. We can also observe in the hopping task that, although only after a considerable number of training iterations and exhibiting a large variance, some torque-actuated runs achieve a larger overall return than the best muscle-actuated runs.

\begin{figure}
\centering
\includegraphics[width=0.6\textwidth]{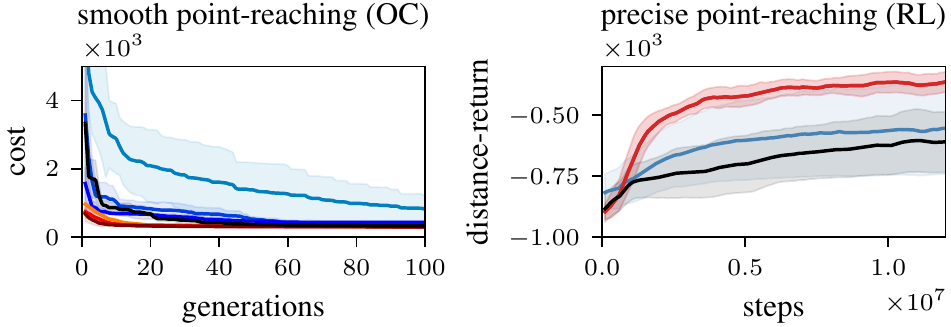}
\\
\definecolor{mycolor1}{rgb}{0.00000,0.50000,0.75000}%
\definecolor{mycolor2}{rgb}{0.55000,0.66250,0.94375}%
\definecolor{mycolor3}{rgb}{0.77500,0.55000,0.55000}%
{\hspace{1.5cm}\tiny\textcolor{mycolor1}{\rule[1.5pt]{8pt}{1pt}} torque, c = 0.05 s \quad \textcolor{blue!50!mycolor1}{\rule[1.5pt]{8pt}{1pt}} torque, c = 0.15 s \quad \textcolor{blue}{\rule[1.5pt]{8pt}{1pt}} torque, c = 0.3 s \quad \textcolor{orange}{\rule[1.5pt]{8pt}{1pt}} muscle, c = 0.05 s \quad
\textcolor{red}{\rule[1.5pt]{8pt}{1pt}} muscle, c = 0.15 s \quad
\textcolor{black!50!red}{\rule[1.5pt]{8pt}{1pt}} muscle, c = 0.3 s} %
{\hspace{1.5cm} \tiny\textcolor{red}{\rule[1.5pt]{8pt}{1pt}} muscle, RL  \quad \textcolor{blue}{\rule[1.5pt]{8pt}{1pt}} torque, RL \quad \textcolor{black}{\rule[1.5pt]{8pt}{1pt}} PD-control, OC and RL}
\vspace{-1.5mm}
\caption{\textbf{Cost value or return in comparison with PD-control for torque. } Left: Muscles outperform all other considered morphologies with OC, while PD-control achieves lower cost than torque actuation with large control resolution $c=0.3$. Right: PD-control does not yield an advantage over torque actuators with RL when applied to the precise point-reaching task.}
\label{fig:pointreaching_pd}
\end{figure}

We additionally investigated a PD controller for the torque actuator morphology, see Fig.~\ref{fig:pointreaching_pd}. 
While the PD controller slightly improves the data-efficiency for some cases, for both OC as well as for RL,  the muscle actuator outperforms all baselines. See \sec{supp:pd} for more experiments.
\newlength\mylinewidth
\setlength\mylinewidth{0.5pt}
\tikzset{
    ultra thin/.style= {line width=0.25\mylinewidth},
    very thin/.style=  {line width=0.5\mylinewidth},
    thin/.style=       {line width=\mylinewidth},
    semithick/.style=  {line width=1.5\mylinewidth},
    thick/.style=      {line width=2\mylinewidth},
    very thick/.style= {line width=3\mylinewidth},
    ultra thick/.style={line width=4\mylinewidth},
    every picture/.style={semithick}
}
\begin{figure}
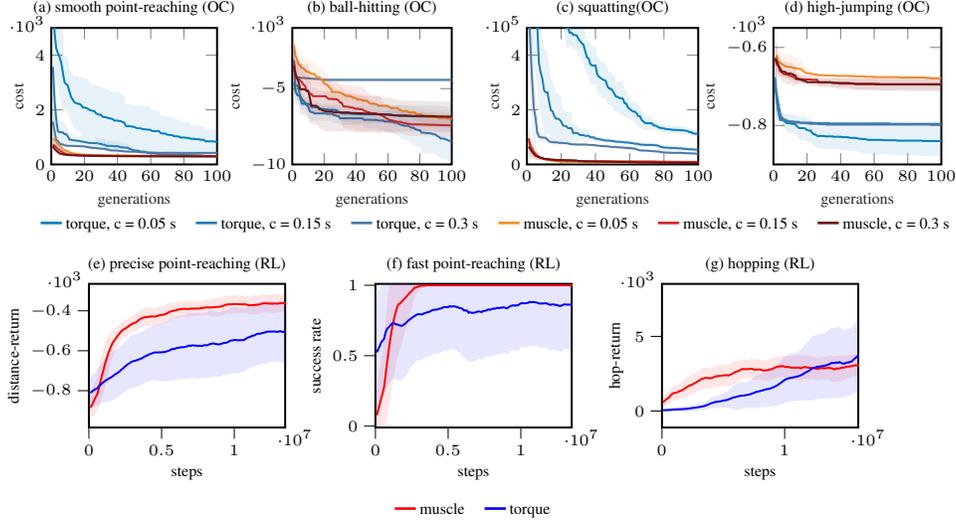

	\tikzsetnextfilename{compare_dataefficiency_all}
	\centering
	\tiny
	\pgfplotsset{
		compat=1.6,
		try min ticks = 3,
		legend image code/.code={
			\draw[mark repeat=2,mark phase=2]
			plot coordinates {
				(0cm,0cm)
				(0.15cm,0cm)        %
				(0.3cm,0cm)         %
			};%
		}
	}
    \definecolor{mycolor1}{rgb}{0.00000,0.50000,0.75000}%
    \definecolor{mycolor2}{rgb}{0.55000,0.66250,0.94375}%
    \definecolor{mycolor3}{rgb}{0.77500,0.55000,0.55000}%
	\definecolor{darkgray176}{RGB}{176,176,176}
\definecolor{darkorange25512714}{rgb}{1.0, 0.0, 0.0}
\definecolor{darkorange}{rgb}{1.0, 0.0, 0.0}
\definecolor{steelblue31119180}{rgb}{0.0,0.0,1.0}
\definecolor{steelblue}{rgb}{0.0,0.0,1.0}
\definecolor{mycolor1}{rgb}{0.00000,0.50000,0.75000}%
\definecolor{mycolor2}{rgb}{0.55000,0.66250,0.94375}%
\definecolor{mycolor3}{rgb}{0.77500,0.55000,0.55000}%
\begin{tikzpicture}
\def\Factor{-3}
\setlength{\figH}{0.13\textwidth}
\setlength{\figW}{0.25\textwidth}
\input{figures/dataefficiency_cost_pointreaching}
\input{figures/dataefficiency_cost_ballhitting}
\input{figures/dataefficiency_cost_squatting}
\input{figures/dataefficiency_cost_jumping}
\setlength{\figH}{0.25\textwidth}
\setlength{\figW}{0.3\textwidth}
\input{figures/arms_precision_random_onlyaxis.tex}
\input{figures/arms_speed_random_onlyaxis.tex}
\input{figures/bipeds_hopping_onlyaxis}
\draw (5.9, -0.8) node {\tiny\textcolor{mycolor1}{\rule[1.5pt]{8pt}{1pt}} torque, c = 0.05 s \quad \textcolor{blue!50!mycolor1}{\rule[1.5pt]{8pt}{1pt}} torque, c = 0.15 s \quad \textcolor{blue}{\rule[1.5pt]{8pt}{1pt}} torque, c = 0.3 s \quad \textcolor{orange}{\rule[1.5pt]{8pt}{1pt}} muscle, c = 0.05 s \quad
\textcolor{red}{\rule[1.5pt]{8pt}{1pt}} muscle, c = 0.15 s \quad
\textcolor{black!50!red}{\rule[1.5pt]{8pt}{1pt}} muscle, c = 0.3 s}; %
\draw (5.6, -4.6) node {\tiny\textcolor{darkorange25512714}{\rule[1.5pt]{8pt}{1pt}} muscle  \quad \textcolor{steelblue31119180}{\rule[1.5pt]{8pt}{1pt}} torque};
\end{tikzpicture}%
    \vspace{-2.2mm}
	\caption{\textbf{Cost value or returns for different tasks.} Plotting the mean and standard deviation (shaded area) for 5 (OC/MPC) or 8 (RL) repeated runs for the two actuator morphologies (muscle in red, torque in blue). Additionally, the temporal control resolution $c$ was varied in the OC cases.}
	\label{fig:data_efficiency}
\end{figure}
\subsection{Robustness to hyperparameter variations}
Tuning a growing number of hyperparameters for learning models typically requires considerable time and computational resources. By analysing hyperparameter sensitivity, we test if tuning with torque or muscle actuator morphologies requires less resources. 

\begin{figure}
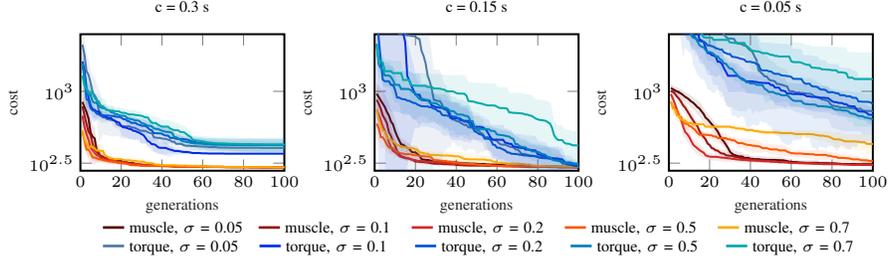

	\tikzsetnextfilename{pointreaching_hyperparameters_sigma}
	\centering
	\tiny
	\pgfplotsset{
		compat=1.6,
		legend image code/.code={
			\draw[mark repeat=2,mark phase=2]
			plot coordinates {
				(0cm,0cm)
				(0.15cm,0cm)        %
				(0.3cm,0cm)         %
			};%
		}
	}
	\setlength{\figH}{0.13\textwidth}
	\setlength{\figW}{0.3\textwidth}
	\definecolor{mycolor1}{rgb}{0.33333,0.00000,0.00000}%
\definecolor{mycolor2}{rgb}{1.00000,0.33333,0.00000}%
\definecolor{mycolor3}{rgb}{1.00000,0.66667,0.00000}%
\definecolor{mycolor4}{rgb}{0.00000,0.16667,0.91667}%
\definecolor{mycolor5}{rgb}{0.00000,0.33333,0.83333}%
\definecolor{mycolor6}{rgb}{0.00000,0.50000,0.75000}%
\definecolor{mycolor7}{rgb}{0.00000,0.66667,0.66667}%
\begin{tikzpicture}
\def\Factor{-3}
\input{figures/arm26_reaching_hyperparameter_sigmaanalysis_c03_onlyaxis.tex}
\input{figures/arm26_reaching_hyperparameter_sigmaanalysis_c015_onlyaxis.tex}
\input{figures/arm26_reaching_hyperparameter_sigmaanalysis_c005_onlyaxis.tex}
\node[text width=10cm] at (4.3,-0.9){\tiny\textcolor{mycolor1}{\rule[1.5pt]{8pt}{1pt}} muscle, $\sigma$ = 0.05\quad \textcolor{red!50!mycolor1}{\rule[1.5pt]{8pt}{1pt}} muscle, $\sigma$ = 0.1 \quad
\textcolor{red}{\rule[1.5pt]{8pt}{1pt}} muscle, $\sigma$ = 0.2 \quad
\textcolor{mycolor2}{\rule[1.5pt]{8pt}{1pt}} muscle, $\sigma$ = 0.5 \quad
\textcolor{mycolor3}{\rule[1.5pt]{8pt}{1pt}} muscle, $\sigma$ = 0.7 \quad
\textcolor{blue}{\rule[1.5pt]{8pt}{1pt}} torque, $\sigma$ = 0.05 \quad
\textcolor{mycolor4}{\rule[1.5pt]{8pt}{1pt}} torque, $\sigma$ = 0.1 \quad \hspace{0.1em}
\textcolor{mycolor5}{\rule[1.5pt]{8pt}{1pt}} torque, $\sigma$ = 0.2 \quad \hspace{0.1em}
\textcolor{mycolor6}{\rule[1.5pt]{8pt}{1pt}} torque, $\sigma$ = 0.5 \quad \hspace{0.1em}
\textcolor{mycolor7}{\rule[1.5pt]{8pt}{1pt}} torque, $\sigma$ = 0.7 }; %
\end{tikzpicture}%
	\vspace{-2mm}
	\caption{\textbf{Muscle morphology is more robust towards hyperparameter variation ($\boldsymbol{\sigma}$) in point reaching.} The cost value of the best observation is shown. The mean and standard deviation (shaded area) are plotted for five repeated runs for the two actuator morphologies (muscle in red, torque in blue) with different control resolutions $c$. }
	\label{fig:sigmaanalysis}
\end{figure}

\Fig{fig:sigmaanalysis} shows the cost curves for smooth point-reaching for the evolutionary optimization algorithm CMA-ES for different values of $\sigma$, which is the principal tuneable parameter for this algorithm. The performance curves vary much more for torque actuators for all considered cases. Furthermore, all muscle-actuated cases find a good-enough optimum with fewer iterations and a smaller variance, independent of the hyperparameter $\sigma$ and the control resolution $c$. 

\begin{figure}
	\tikzsetnextfilename{pointreaching_hyperparameter_tpred}
	\centering
	\tiny
	\pgfplotsset{
		compat=1.11,
		legend image code/.code={
			\draw[mark repeat=2,mark phase=2]
			plot coordinates {
				(0cm,0cm)
				(0.15cm,0cm)        %
				(0.3cm,0cm)         %
			};%
		}
	}
	\setlength{\figH}{0.11\textwidth}
	\setlength{\figW}{0.32\textwidth}
	\input{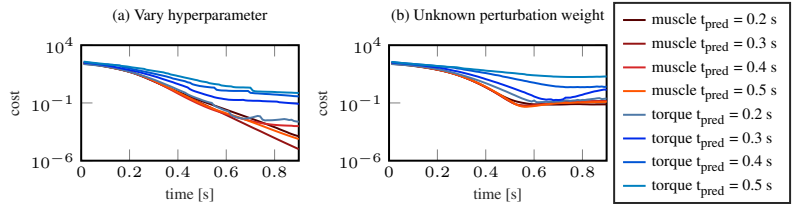}
	\caption{\textbf{Muscle morphology is more robust towards variation of hyperparameter $\mathbf{t}_\text{pred}$ and unknown perturbations.}  Plotting the development of cost over time for the two actuator morphologies (muscle in red, torque in blue) while varying the hyperparameter $t_\text{pred}$ denoting the length of the prediction horizon. Left: The unperturbed case. Right: The prediction model is not aware of the added weight to the lower arm.}
	\label{fig:mpc_pred_horizon}
\end{figure}

The same task was repeated using MPC while varying the main hyperparameter $t_\text{pred}$, which represents the prediction horizon in moving horizon strategies. The performance curves and the final cost vary much more for torque actuators (\fig{fig:mpc_pred_horizon}a, note, the cost is plotted logarithmically).  

\begin{figure}
\centering
\includegraphics[width=1.0\textwidth]{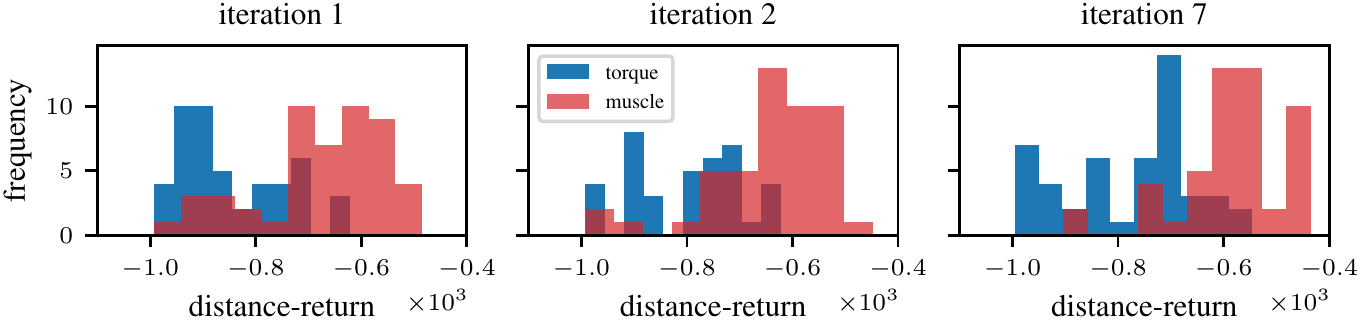}
\vspace{-0.65cm}
	\caption{\textbf{Muscle actuators need less parameter tuning for good performance.} Hyperparameters are optimized for precise point-reaching following an iterative sampling scheme, each run trains for $2\times 10^{6}$ iterations. Fifty sets of parameters are sampled randomly from pre-determined distributions, the final performance is evaluated and used to adapt the sampling distributions for the next iteration. We plot the return distributions over the sampled parameters at different iterations.}
\label{fig:hyperpararl}
\end{figure}

Finally, we performed an extensive hyperparameter optimization for precise point-reaching. For each iteration, $50$ sets of parameters are randomly chosen and the final task performance is evaluated after $2 \times 10^{6}$ learning iterations. The sampling distributions for the parameters are then fit to the best performing runs and $50$ additional sets are evaluated for the next iteration. We optimize the learning rates of MPO, as well as gradient-clipping thresholds, as these have a strong influence on learning speed and stability. Muscle actuators already outperform torque-actuators in the first iteration, with a greater number of well performing parameter sets (\fig{fig:hyperpararl}). Almost no low-performing runs remain for iteration $7$, while a large torque-performance is only achieved by a small subset of parameter settings. See \supp{supp:additional} for more hyperparameter variations.
\subsection{Robustness to perturbations}
\begin{figure}
\centering
\includegraphics[width=0.49\textwidth]{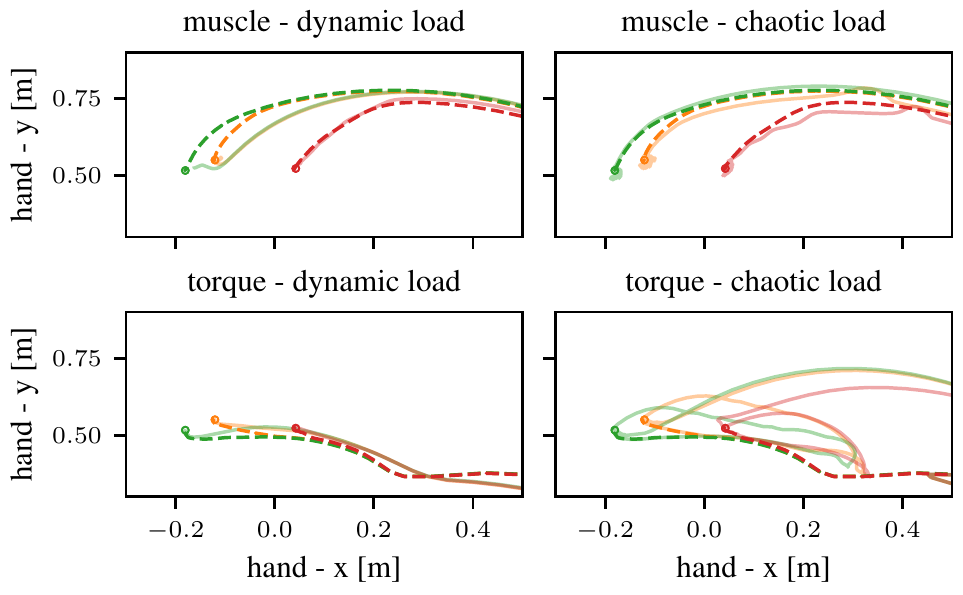}
\includegraphics[width=0.49\textwidth]{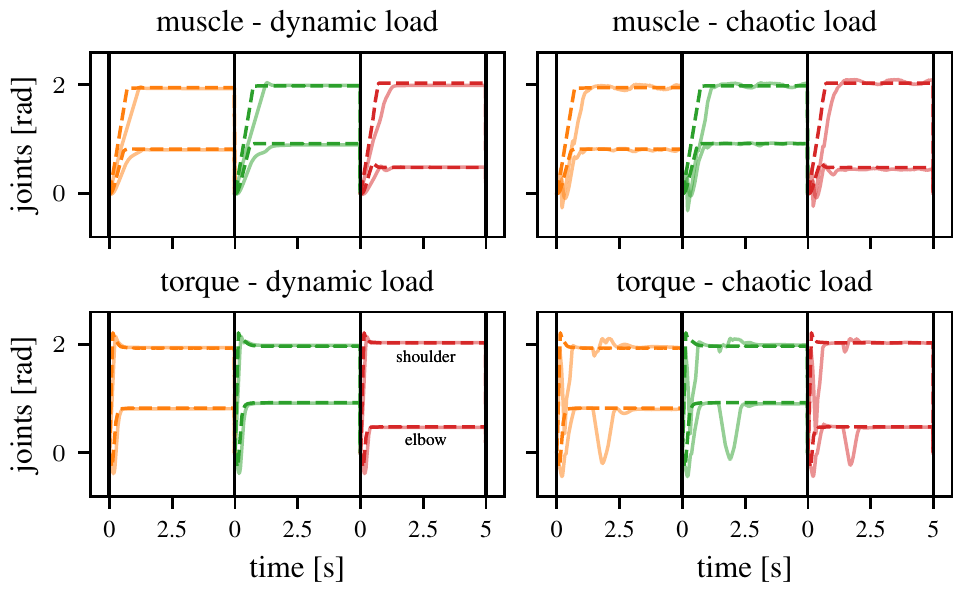}
\vspace{-0.2cm}
	\caption{\textbf{Trajectories for dynamic ($\mathbf{1.5}$ kg weight) and chaotic (attached ball) load.} Left: Three random goals are exemplarily shown, the respective goal position is marked as a circle. The unperturbed baseline for each goal is shown with a dashed line. Right: Joint trajectories for the same experiment, the unperturbed baselines are shown with dashed lines. Vertical bars mark episode resets.}
\label{fig:rl_robustness_arm}
\end{figure}

In this section, we probe the robustness of the obtained policies against unknown perturbations. In precise point-reaching, we evaluated the RL reaching policies for two modifications that were not present during training: First, the hand-weight of the model is increased by $1.5$ kg (dynamic load), and secondly a free spherical weight is attached to the end effector with a cable (chaotic load). We can see in \fig{fig:rl_robustness_arm} that the muscle-based policy does not suffer significant changes in performance, except for a small circular motion ($3$ cm) around the goal position in the chaotic load case. In contrast, the torque actuator morphology leads to unstable reaching and strong oscillations. Both morphologies seem to handle the dynamic load well. See \supp{supp:additional} for more goals.

For the MPC controller, we evaluated robustness by introducing environment changes that are unknown to the prediction model. One example is the lifting of an object with unknown weight, a typical robotics task.  When adding $1$ kg to the lower arm of the \armdemoa model (\fig{fig:mpc_pred_horizon}), the performance in both actuator cases is worse than in the unperturbed case (left); the movement is also slower. However, the variance and absolute value of the final cost in the muscle-actuated case are still much lower compared to the torque-actuated case (plotted logarithmically). See Suppl. \ref{sec:mpc_more_weights} for more weight variations. \looseness=-1

\begin{figure}
	\tikzsetnextfilename{squatting_perturbation_unknownforce}
	\centering
	\tiny
	\pgfplotsset{
		compat=1.11,
		legend image code/.code={
			\draw[mark repeat=2,mark phase=2]
			plot coordinates {
				(0cm,0cm)
				(0.15cm,0cm)        %
				(0.3cm,0cm)         %
			};%
		}
	}
	\setlength{\figH}{0.13\textwidth}
	\setlength{\figW}{0.36\textwidth}
	\input{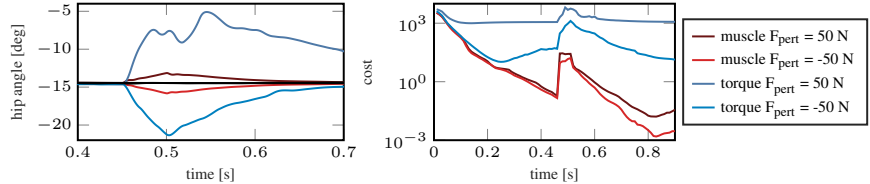}
	\caption{\textbf{Hip and cost trajectory for squatting with unknown perturbation forces.}
	The muscle morphology is shown in red, the torque morphology in blue while varying the perturbation force $F_\text{pert}$ [N] (applied between $t>=0.45$ s and $t<=0.5$ s).}
	\label{fig:squat_mpc_force}
\end{figure}

\begin{figure}
\centering
\begin{subfigure}[b]{0.51\textwidth}
\centering
{\hspace{0.75cm}\small\textcolor{red}{\rule[1.5pt]{8pt}{1pt}} muscle \quad \textcolor{ourblue}{\rule[1.5pt]{8pt}{1pt}} torque}\\
\includegraphics[width=0.95\textwidth]{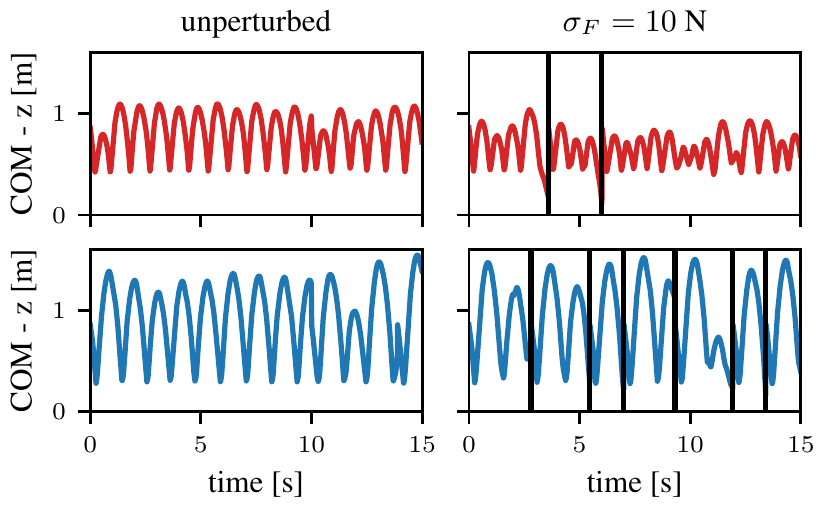}
\end{subfigure}
\hspace{1cm}
\begin{subfigure}[b]{0.301\textwidth}
\vspace{-5cm}
\includegraphics[width=0.95\textwidth]{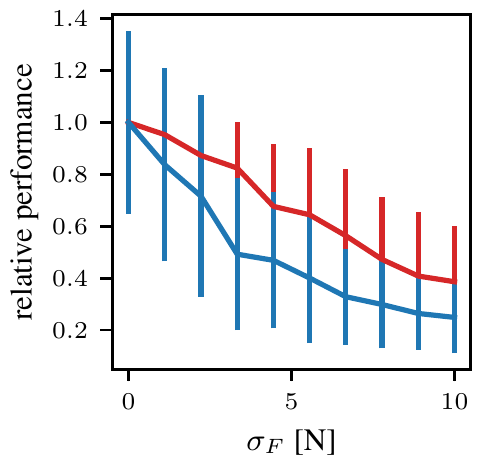}
\end{subfigure}
\vspace{-0.25cm}
	\caption{\textbf{Muscle actuators lead to more robust hopping.} At each time step, there is a 5 \% chance of a random force being applied to the hip, knee, and ankle joints. Left: COM motion over time. Vertical bars mark episode resets due to falls of the \biped. Right: Relative performance for different standard deviations of the random force $\sigma_{F}$. Performance is scaled by the unperturbed mean return.}
\label{fig:rl_robustness_biped}
\end{figure}

For periodic hopping with the \biped model, we evaluated trained RL policies with random forces that were drawn from a Gaussian distribution $F\sim\mathcal{N}(\cdot|0,\sigma_{F})$ and applied to the hip, knee, and ankle joints with a probability of $0.05$ at each time step. We see in \fig{fig:rl_robustness_biped} that the torque actuator morphology is stronger affected in relative performance than the muscle morphology. In the robustness investigation with MPC in the \allmin squatting task, a force is applied to the hip joint after the system has reached its desired position. \Fig{fig:squat_mpc_force} (left) shows that the desired joint angles are much less affected by the perturbation when muscle actuators are controlled. Furthermore, the cost value associated with the movement recovers much slower for torque actuators (\fig{fig:squat_mpc_force} right).

\section{Discussion}
\label{sec:discussion}
We investigated if muscle-like actuators have beneficial effects for modern learning methods in terms of data-efficiency, hyperparameter sensitivity  and robustness to perturbations. A multitude of variations across physics simulators, learning algorithms and task domains was considered in order to showcase the potential of the considered morphologies independently of the underlying implementation. We showed that muscles yield benefits in tasks requiring stable motion, even when compared to idealized torque actuators, which can be considered an upper performance bound. Indeed, the used torque actuators are able to instantaneously output any desired force at any point of the trajectory, while muscles only slowly change their output due to activation dynamics and can only produce kinematics-dependent force output. 
Despite these limitations, the considered learning algorithms learn more efficiently with muscle actuation in all tasks, except for extreme motions where objectives require a strong force application without stability considerations, such as ball-hitting and high-jumping. In bipedal hopping, it was found that muscles result in more efficient learning, even though some torque-runs achieve higher asymptotic performance. Finally, we observe muscle actuation to result in increased robustness to perturbations and hyperparameter variations, which can facilitate learning on real robotic systems that not only present sensor and motor noise, but also prohibit extensive parameter searches.\looseness=-1
\paragraph{Outlook for real-world robotics}
We see two use-cases of our findings: (1) Muscular force-length-velocity and low-pass filter characteristics can be implemented as low-level actuator control for torque-controlled robotic systems (e.g., \citep{GarciaCordova2001a, Seyfarth2007, Ijspeert2020a, rai2018bayesian}). This could allow us to exploit the improved data efficiency and robustness observed in our study for RL on a real robotic system. (2) Novel soft robotic actuators, such as artificial muscles~\citep{boblan2007human,klute1999mckibben,vanderborght2013variable,wolfen2018bioinspired}, promise to revolutionize specific application scenarios of robotics, e.g., wearable rehabilitation devices \citep{Zhu2022a}. While soft actuated systems are hard to control from a classical control theory point of view, our results and other works \citep{peng2017learning} suggest that RL may even benefit from their properties. 
In our study, the simplified MuJoCo muscle model is applicable as a low-level controller in the sense of the first use case, while the results with the complex series-elastic muscle model in Demoa highlights the second use-case, making both cases strong arguments to consider RL and muscle properties a promising combination.

\paragraph{Limitations}
Although we have reported results for a wide variety of algorithms and tasks, we cannot give theoretical statements about the general applicability of our findings. Additionally, some of the tasks we employed were limited in complexity and might also be solvable with classical control algorithms. The MuJoCo muscle model, while computationally efficient, only captures rudimentary properties of biological systems. The demoa implementation, on the other hand, includes visco-elastic, passive tendon characteristics and muscle routing as joint angle-dependent lever arms to account for many physiological details---at substantial additional computational cost. Finally, learning with intermediate control signals given to impedance or position controllers, instead of direct torque commands, might also improve learning performance, while muscle-like properties could have been introduced by learning priors or additional cost terms.

\clearpage
\acknowledgments{We thank Daniel Höglinger for help during the development of the hopping reward function. Furthermore, we like to thank Marc Toussaint, Danny Driess and David Holzm\"{u}ller for initial discussions regarding the topic of learning with muscles. This work was supported by the Deutsche Forschungsgemeinschaft (DFG, German Research Foundation) under Germany’s Excellence Strategy - EXC 2075 - 390740016 (SimTech). We thank the International Max Planck Research School for Intelligent Systems (IMPRS-IS)
for supporting all authors. Georg Martius is a member of the Machine Learning Cluster of Excellence, EXC number 2064/1 – Project number 390727645.
This work was supported by the Cyber Valley Research Fund (CyVy-RF-2020-11 to DH and GM).
We acknowledge the support from the German Federal Ministry of Education and Research (BMBF) through the Tübingen AI Center (FKZ: 01IS18039B).}

\bibliography{literature}  %
\newpage

\begin{center}\Large \textbf{Supplementary Material for \\
Learning with Muscles: Benefits for Data-Efficiency and Robustness in Anthropomorphic Tasks}\end{center}
\section{Muscle model}
In this section, the implementations of the muscle models for demoa and MuJoCo are described. The demoa model approximates biological muscles with more physiological detail and accuracy, whereas the simpler MuJoCo model allows the simulation of rudimentary muscular properties at minimal computational cost, rendering it usable for machine learning. 
\label{supp:muscles}
\subsection{Muscle model in Mujoco} 
Even though the MuJoCo simulator includes the capability of simulating muscles, it requires the explicit definition of tendon insertion points and wrapping surfaces for each model. We, therefore, use our own muscle implementation for the MuJoCo experiments, that does not contain tendons. As a direct consequence, the muscle-fiber length is \textit{uniquely} determined by the joint angle.

In the following, we describe activation dynamics, definitions of muscle-fiber length and velocity, the computation of the resulting torque and the parametrization. 

\begin{figure}
    \centering
    \includegraphics[width=1.0\textwidth]{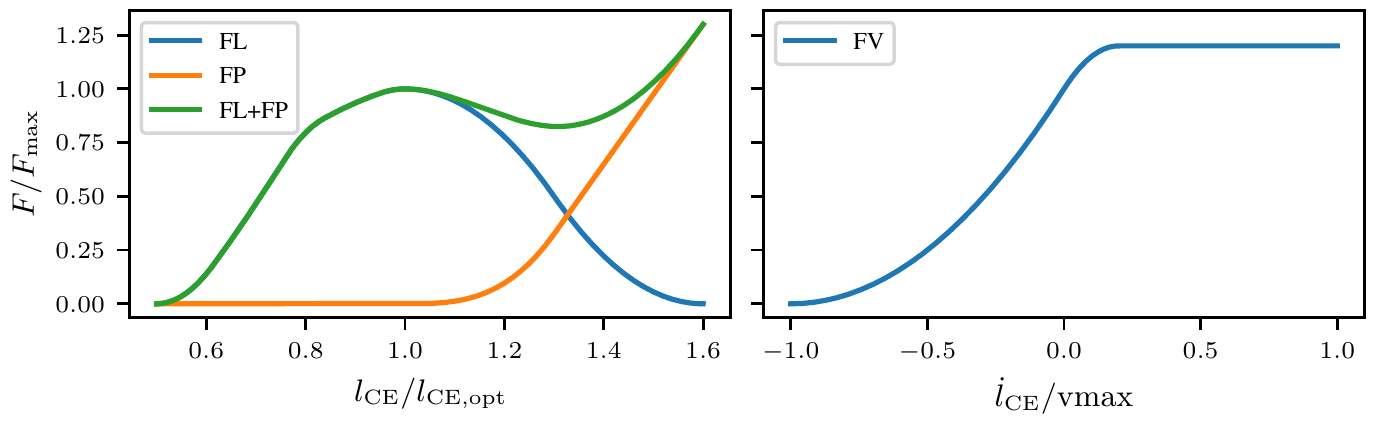}
    \caption{Force-length (FL) and force-velocity (FV) relationships and passive force (FP) used in MuJoCo~\cite{todorovMuJoCoPhysicsEngine2012a}. We use the same phenomenological functions in our own MuJoCo muscle model.  While FL and FV get scaled by the current muscle activity $a$, FP does not (see \eqn{eq:forcemujoco}).}
    \label{fig:musclemodelmujoco}
\end{figure}

\paragraph{Muscle-tendon-unit}
Each controllable joint of the MuJoCo model is actuated by two monoarticular muscles and we do \textbf{not} compute tendon length.
We assume that:
\begin{equation}
l_{\mathrm{MTU}} = l_{\mathrm{CE}}, 
\label{eq:length}
\end{equation}
where $l_{\mathrm{MTU}}$ is the length of the entire muscle-tendon-unit and $l_{\mathrm{CE}}$ is the length of the muscle fiber, or contractile element.
We define muscle-fiber length and velocity by a linear equation~\cite{Kistemaker2006, Geyer2010, Bayer2017}:
\begin{align}
    l_{\mathrm{CE},i} &= m_{i}\, \phi_{j} + l_{\mathrm{ref},i}\\
    \dot{l}_{\mathrm{CE},i} &= m_{i}\, \dot{\phi_{j}},
\end{align}
where $\phi_{j}$ is the joint angle, $m_{i}$ and $l_{\mathrm{ref},i}$ are computed from user-defined parameters, and $i\in\{1, 2\}$, as we assume two antagonistic muscles per joint. The parameter $m_{i}$ acts as a constant moment arm in our model, see \eqn{eq:torquemujoco}.
\paragraph{Activation dynamics}
The evolution of muscle activity obeys the following first-order ordinary differential equation:
\begin{equation}
    \dot{a}(t) = \frac{1}{\Delta t_{a}} (u(t) - a(t)), 
    \label{eq:act_dyn_mujoco}
\end{equation}
where $u(t)$ is a control signal.
\paragraph{Muscle force}
Given the previous quantities, the muscle force is computed by:
\begin{equation}
F_{i} = \left[ \mathrm{FL}(l_{\mathrm{CE},i}) \, \mathrm{FV}(v_{\mathrm{scale}}\,\dot{l}_{\mathrm{CE},i}) a_{i} + \mathrm{FP}(l_{\mathrm{CE},i})\right] \, F_{\mathrm{max},}    
\label{eq:forcemujoco}
\end{equation}\\
where $v_{\mathrm{scale}}$ is a scaling parameter to adjust in which region of the force-velocity (FV)-curve typical fiber velocities operate.
We can then compute the resulting joint torque:
\begin{equation}
    \tau = - (m_{1}\, F_{1} + m_{2}\, F_{2}).
    \label{eq:torquemujoco}
\end{equation}
The functions FL, FV and FP are given by MuJoCo internal functions that phenomenologically model experimental data and are applied to normalized muscle lengths and velocities, see MuJoCo documentation~\cite{todorovMuJoCoPhysicsEngine2012a} and \fig{fig:musclemodelmujoco}.
\paragraph{Parametrization} As the there is a one-to-one mapping of joint angle to muscle lengths in our model, we can determine the required parameters $m_{i}$ and $l_{\mathrm{ref},i}$, if a mapping of $l_{\mathrm{min}}$ to $\phi_{\mathrm{min}}$ and $l_{\mathrm{max}}$ to $\phi_{\mathrm{max}}$ is specified (assuming $l_{\mathrm{max}}$ to be the maximal muscle-fiber length $l_{\mathrm{CE}}$). Inserting them into \eqn{eq:length} and solving the resulting system of equations gives:
\begin{align}
m_{1} &= \frac{l^{\mathrm{max}} - l^{\mathrm{min}}}{\phi^{\mathrm{max}} - \phi^{\mathrm{min}} + \epsilon}\\
l^{\mathrm{ref}}_{1} &= l_{\mathrm{min}} - m_{1}\, \phi_{\mathrm{min}}\\
m_{2} &= \frac{l^{\mathrm{max}} - l^{\mathrm{min}}}{\phi^{\mathrm{min}} - \phi^{\mathrm{max}} + \epsilon} \\
l^{\mathrm{ref}}_{2} &= l_{\mathrm{min}} - m_{2}\,\phi_{\mathrm{max}} \\
\end{align}
All in all, $\phi^{\mathrm{min}}$, $\phi^{\mathrm{max}}$, $l^{\mathrm{min}}$, $l^{\mathrm{max}}$ and $F_{\mathrm{max}}$ are required to be specified. The constant $\epsilon=0.01$ ensures numerical stability. We use the same parametrization for each MuJoCo task, see  \tab{tab:mujocomuscle}.

The maximum and minimum joint angles were chosen to allow for a large range of motion.  They do not constitute hard limits, but the passive elastic force FP will increase strongly when reaching them. The maximum and minimum fiber lengths are identical to the MuJoCo default values. As we want to study the benefits of muscular properties in learning, we chose the time and velocity scales $\Delta t_{a} = 0.01$ and $v_{\mathrm{scale}}=0.5$ to be large enough to produce noticeable effects, such as low-pass filtering and self-stabilization properties, across all performed tasks. To determine maximum muscle forces, we trained muscle-actuator policies for a chosen maximum force value, after which we adjusted maximum torque-actuator forces to be identical or slightly larger to the maximally observed muscle forces in the final task policies. We repeated this procedure with different force values until good performance could be observed for both morphologies, see \supp{supp:forcevariation} for an evaluation across different force values. All other MuJoCo internal parameters related to muscle modeling are kept to their default values.

In practice, we implement the muscle model in Cython~\cite{behnel2011cython}, interfacing with OpenAI gym~\cite{gym}, which achieves similar execution speed to native MuJoCo. 
\begin{table}
\caption{Parameters for the MuJoCo muscle morphology.}
\centering
\begin{subtable}[t]{.49\textwidth}
\caption{Muscle parameters}
\centering
\begin{tabular}{@{}ll@{}}
\textbf{Parameter} & \textbf{Value}  \\
 \toprule
 $l_{\mathrm{max}}$ & 1.05 \\
 $l_{\mathrm{min}}$ & 0.75 \\
 $\phi_{\mathrm{max}}$ & $\pi/2$ [rad]\\
 $\phi_{\mathrm{max}}$ & -$\pi/2$ [rad]\\
 $\Delta t_{a}$ & 0.01 [s]\\
 $v_{\mathrm{scale}}$ & 0.5 \\
\bottomrule
\end{tabular}
\end{subtable}
\begin{subtable}[t]{.49\textwidth}
\caption{Maximum isometric force}
\centering
\begin{tabular}{@{}ll@{}}
\textbf{Task} & \textbf{Value}  \\
 \toprule
 \armmujoco & 295 [N]\\
 \biped & 5000 [N]\\
\bottomrule
\end{tabular}
\end{subtable}
\label{tab:mujocomuscle}
\end{table}

\subsection{Muscle model in demoa}
\label{sec:muscledemoa}
\begin{figure}
    \centering
	\includegraphics[height=2.5cm]{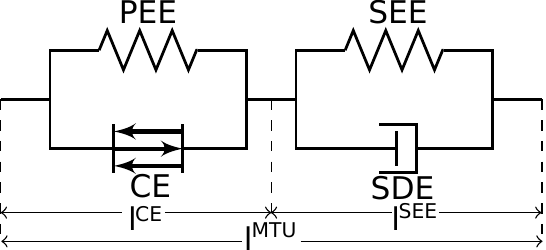}
	\caption{The muscle model in demoa is modeled as lumped Hill-type muscle model (figure adapted from \citet{haeufle2014hill}).  \label{fig:hilltype}}
\end{figure}

The muscle model implemented in demoa \citep{darus-2550_2022} includes additionally visco-elastic, passive tendon characteristics and muscle routing as joint angle-dependent lever arms to account for many physiological details. In the following, we describe the activation and contraction dynamics of the muscle model, as well as the tendon characteristics and the nonlinear lever arms. 
\paragraph{Activation dynamics}
The muscles are activated with the learned and optimized control signal $u$, which is nonlinearly transformed into an activation signal. The activity $a$ is following a first-order differential equation of normalized calcium ion concentration $\gamma$ as introduced by \citet{hatze1977myocybernetic} and simplified by Rockenfeller et al. \citep{Rockenfeller2015,Rockenfeller2018}:
\begin{equation}
    \dot{\gamma}(t)=M_{\mathrm{H}}(\stim(t)-\gamma(t)) \label{eq:actdyn1}
\end{equation}
and a nonlinear mapping onto the muscles activity
\begin{equation}
\act(t)=\frac{\act_0+\varpi}{1+\varpi}, \label{eq:actdyn2}
\end{equation}
with $\varpi(\gamma(t), \lce(t))=(\gamma(t)\cdot\rho(\lce))^\nu$ and $\rho(\lce)=\varpi_\mathrm{opt}\cdot\tfrac{\lce}{l_\text{opt}}=\gamma_\mathrm{c}\cdot\rho_0\cdot\tfrac{\lce}{l_\mathrm{opt}}$. The parameter values are chosen muscle non-specifically and are given in the description of the models (see \citep{darus-2871_2022,darus-2982_2022}). 
\paragraph{Muscle-tendon-unit}
The predicted forces are modeled using Hill-type muscle models \citep{haeufle2014hill} including four spring-damper components (see Fig. \ref{fig:hilltype}): The contractile element (CE) models the active force production of biological muscle fibers, including the nonlinear \textit{force-length} and nonlinear \textit{force-velocity} relation. The parallel elastic element (PEE) models the passive connective tissue in the muscle belly and is arranged in parallel to the CE. The visco-elastic properties of the tendons are modeled using a serial elastic element (SEE) and a serial damping element (SDE). 
All in all, the governing model dependencies for all muscles $i=1,...,n$ are: 
\begin{eqnarray}
\dlcei &=& \fce (\lcei, \lmtui, \dlmtui, \act_i) \label{eq:contractdyn}\\
\dot \act_i &=& f_\act (\act_i, \stim_i, \lcei) \label{eq:actdyn3}\\
\fmtui &=& \fmtui(\lmtui, \dlmtui, \lcei, \act_i) \label{eq:forcemusc}\,,
\end{eqnarray}
where the first differential equation (Eq. \ref{eq:contractdyn}) denotes the contraction dynamics which models the velocity $\dlce$ of the contractile element. %
This contraction velocity is dependent on the current CE length $\lce$, the length and contraction velocity of the muscle-tendon unit $\lmtu$ and $\dlmtu $ respectively, and the activity $a$. The latter is modeled by the activation dynamics (see Eq. \ref{eq:actdyn1},\ref{eq:actdyn2},\ref{eq:actdyn3}). Finally, a force $\fmtui$  for each muscle is produced which is translated into joint torques. 
\paragraph{Nonlinear lever arms}
To translate the force into joint torques, the muscle path around the joints is routed via deflection ellipses in demoa \citep{hammer2019tailoring}. If the length of the half-axises of all ellipses are set to zero, this approach can be simplified to the more commonly used fixed via-point approach for muscle routing. Based on the resulting moment arms of the muscles, the force $\fmtu$ is translated to generalized torques acting on the degrees of freedom of the system. 
\paragraph{State of the system}
Using a musculoskeletal model with a Hill-type muscle model, as described in this section, increases the number of state variables because two additional differential equations need to be solved for each included muscle. The entire state vector $x$ can therefore be formulated as: 
\begin{equation}
    x \in \mathbb{R}^{2n_\text{musc}+ 2n_\theta} = \{\gamma_i, l_{\text{CE},i}, \theta_j, \dot{\theta}_j\} 
\end{equation}
where $\theta$ and $\dot{\theta}$ represent the generalized joint angle coordinates and their respective velocities, and $n_\text{musc}$ and $n_\theta$ denote the number of muscles and the number of joints, respectively. 
\section{Experimental details and hyperparameters}
In this section, we describe algorithm implementation details while also reporting additional settings that were used to obtain previously shown results, as well as used hyperparameters. The section is divided such that experiment details are shown with the control algorithm that was used to generate the results.
\label{supp:implementation}
\subsection{Optimal Control (OC)}
\label{supp:optimal_control}
In the optimal control case, we used the covariance matrix adaptation evolution strategy (CMA-ES) \citep{hansen2003reducing} to find the control policy $u(k)$. As mentioned in the main paper, we chose the same population size and number of generations for both actuator morphologies to allow for a fair comparison, even though the number of decision variables $n_u$ is always larger in the muscle-actuated case. In all cases, if not otherwise mentioned, we use a fixed population size of $36$ and a fixed number of generations of $100$ while varying the control resolution $c$. If the control resolution is refined, this correlates to an increase in the number of decision variables $n_u$, however, we specifically did not change the population size or generation number because we wanted to compare the data-efficiency and learning with limited resources for the chosen actuators. 
The temporal control resolution $c$ was typically varied for $c = \{0.05, 0.15, 0.3\}$ s. The upper bound of these control resolutions ($c = 0.3$ s) corresponds to a triphasic control pattern for a typical movement duration of $0.9$ s as it was selected in the smooth point-reaching and squatting task. This selection of $c$ was inspired by biological experiments, where it was shown that triphasic patterns occur in muscle surface electromyograms in typical point-reaching movements (e.g. see \citep{wierzbicka1986role,kistemaker2006equilibrium}). The main hyperparameter of the CMA-ES algorithm $\sigma$ was set to the default value of $0.2$ if not otherwise stated.
\subsection{Model Predictive Control {MPC}}
We employed a warm start procedure using the CMA-ES optimizer and afterwards started the MPC routine with a local optimizer BOBYQA \citep{powell2009bobyqa} (part of the standard python optimization package NLOPT). As temporal control resolution in this closed-loop setting, we chose a very fine resolution of $c = 0.01$ s, similar to the RL setup. This allows counteracting perturbations. The prediction horizon was varied between $t_\text{pred} = \{0.2, 0.3, 0.4, 0.5\}$ s as shown in the result section of the main paper.    
\subsection{Reinforcement Learning (RL)}
\label{supp:rl}
We use the RL algorithm MPO~\cite{abdolmaleki2018maximum}, implemented in TonicRL~\cite{pardo2020tonic}. Hyperparameters were optimized with a simplified in-house CEM optimizer. All RL experiments are averaged over $8$ random seeds except for the hyperparameter optimization, which would have been computationally intractable. Each experimental run was computed with 1 NVIDIA V100 GPU and 20 CPUs of varying speed and type. We use a fixed control resolution of $c = 0.01$ s for all RL experiments.
\subsubsection{Experimental details}
We give further experimental details in this section. 
\label{supp:expdetails}
\paragraph{Data-efficiency} For the point-reaching experiments, we used the hyperparameters that were found in the meta-optimization (see \fig{fig:hyperpararl}) for both morphologies. For the hopping task, we used default MPO parameters. The updated learning curves with optimized parameters, as well as additional results on hyperparameters and maximum force settings can be found in \supp{supp:additional}.  

\paragraph{Hyperparameter optimization} We optimized the performance of both actuator morphologies in the precise point-reaching task in MuJoCo (see \fig{fig:supphyperpararlarm}). For each iteration, $N_{\mathrm{sets}}$ sets of random parameters are drawn from fixed normal and log-normal distributions. For each of these sets, the task performance is evaluated after $T_{\mathrm{train}}$ environment interactions, where $T_{\mathrm{train}}$ is chosen such that a noticeable increase in performance can be observed with both actuator morphologies. After each iteration, $M_{\mathrm{elite}}$ elite parameter sets are chosen and the mean and standard deviation of each parameter-generating distribution is updated by fitting a (log-)normal distribution to the $M_{\mathrm{elite}}$ elite sets with maximum-likelihood estimation. See \tab{tab:hyperpara} for exact specifications. The meta-optimization for hopping can be found in \fig{fig:supphyperpararlbiped}.

In the experiments, we only used truncated log-normal distributions to generate parameters. The samples were clipped to the bounds given in \tab{tab:hyperpara} and initial mean and standard deviation were chosen to lie inside the bounded interval. More precisely, we defined $\log(\mu) = (a + b)/2$ and $\log(\sigma) = (b - a)/4$, where $a$ and $b$ are the chosen bounds. The chosen parameters were the actor learning rate $\mathrm{lr}_{\mathrm{a}}$, the critic learning rate $\mathrm{lr}_{\mathrm{c}}$, the learning rate of the dual optimizer $\mathrm{lr}_{\mathrm{d}}$, the gradient clipping threshold for the actor $\mathrm{clip}_{\mathrm{a}}$ and the critic $\mathrm{clip}_{\mathrm{a}}$.

\paragraph{Robustness point-reaching} We trained policies with both morphologies in precise point-reaching for $1.5\times 10^{7}$ iterations. The best performing policies were then evaluated for the perturbation experiments. For dynamic load, the mass of the hand is increased by $1.5$ kg to simulate an object. For chaotic load, a ball with radius $0.12$ m and a density of $1000$ kg/m$^{3}$ is attached to a cable of length $0.6$ m, that is connected to the hand. We sample $10$ random goals from the training distribution and visualize three trajectories such that there are no overlapping paths. All $10$ goals are shown in \fig{fig:suppmoregoals}.

\paragraph{Robustness hopping} We trained policies for both morphologies for hopping with the hyperparameters obtained in \fig{fig:supphyperpararlbiped} and for $1.5\times 10^{7}$ iterations. We then record $100$ evaluation episodes where random forces drawn from $F_{i} \sim \mathcal{N}(\cdot|0, \sigma_{F})$ are applied with a probability of $p=0.05$ to the hip, knee and ankle joints and to the pelvis position and rotation. Center of mass trajectories are shown for an interval of $15$ s in the main manuscript, black vertical bars mark episode resets due to extreme angles of the biped, which would cause it to fall to the ground. The performance for each perturbation level is divided by the unperturbed performance for each morphology to yield a relative performance comparison.
\begin{table}
\centering
\caption{Settings for the hyperparameter search.}
\begin{subtable}[t]{.49\textwidth}
\caption{Precise point-reaching}
\centering
\begin{tabular}{@{}lll@{}}
& \textbf{Parameter} & \textbf{Value}  \\
 \toprule
 & $N_{\mathrm{sets}}$ & $50$\\
 & $T_{\mathrm{train}}$ & $2\times 10^{6}$\\
 & $M_{\mathrm{elite}}$ & $10$ \\
\bottomrule
\end{tabular}
\end{subtable}
\begin{subtable}[t]{.49\textwidth}
\caption{Hopping}
\centering
\begin{tabular}{@{}lll@{}}
& \textbf{Parameter} & \textbf{Value}  \\
 \toprule
 & $N_{\mathrm{sets}}$ & $20$\\
 & $T_{\mathrm{train}}$ & $5\times 10^{6}$\\
 & $M_{\mathrm{elite}}$ & $10$ \\
\bottomrule
\end{tabular}
\end{subtable}\\
\begin{subtable}[t]{.49\textwidth}
\caption{Initial distributions}
\centering
\begin{tabular}{@{}lll@{}}
\textbf{Parameter} & \textbf{Distribution} & \textbf{Bounds}  \\
 \toprule
 $\mathrm{lr}_{\mathrm{a}}$ & truncated log-normal & $[2.5\times 10^{-4}, 3\times 10^{-2}]$ \\
 $\mathrm{lr}_{\mathrm{c}}$ & truncated log-normal & $[0.5\times 10^{-2}, 10^{-1}]$ \\
 $\mathrm{lr}_{\mathrm{d}}$ & truncated log-normal & $[0.5\times 10^{-2}, 1]$ \\
 $\mathrm{clip}_{\mathrm{a}}$ & truncated log-normal & $[10^{-7}, 10^{-4}]$ \\
 $\mathrm{clip}_{\mathrm{c}}$ & truncated log-normal & $[10^{-7}, 10^{-4}]$ \\
\bottomrule
\end{tabular}
\end{subtable}

\label{tab:hyperpara}
\end{table}
\subsubsection{Hyperparameters}
The hyperparameters for all RL tasks were set to the best performing runs in the shown hyperparameter optimization. They best trained policies were then used for the perturbation experiments.
\begin{table*}
    \centering
     \caption{RL parameters for MPO in TonicRL for the different tasks. Non-reported values are left to their default setting in TonicRL~\cite{pardo2020tonic}. Common MPO settings are equal for all experiments.}
    \label{tab:hyperparaused}
    \begin{subtable}[t]{.5\textwidth}
        \centering
        \caption{Point-reaching MPO (muscle)}
        \begin{tabular}{@{}lll@{}}
        \toprule
        &\textbf{Parameter} & \textbf{Value} \\
        \midrule
        & $\mathrm{lr}_{\mathrm{a}}$ & $3\times 10^{-4}$ \\
        & $\mathrm{lr}_{\mathrm{c}}$ & $10^{-3}$ \\
        & $\mathrm{lr}_{\mathrm{d}}$ & $2\times 10^{-2}$ \\
        & $\mathrm{clip}_{\mathrm{a}}$ & $4 \times 10^{-5}$ \\
        & $\mathrm{clip}_{\mathrm{c}}$ & $3 \times 10^{-5}$ \\
        & batch size & $100$ \\
        & return-normalizer & No \\
        \bottomrule
        \end{tabular}
    \end{subtable}%
    \begin{subtable}[t]{.5\textwidth}
        \centering
        \caption{Point-reaching MPO (torque)}
        \begin{tabular}{@{}lll@{}}
        \toprule
        &\textbf{Parameter} & \textbf{Value} \\
        \midrule
              & $\mathrm{lr}_{\mathrm{a}}$ & $10^{-3}$ \\
        & $\mathrm{lr}_{\mathrm{c}}$ & $5\times 10^{-3}$ \\
        & $\mathrm{lr}_{\mathrm{d}}$ & $8.2 \times 10^{-3}$ \\
        & $\mathrm{clip}_{\mathrm{a}}$ & $7 \times 10^{-6}$ \\
        & $\mathrm{clip}_{\mathrm{c}}$ & $10^{-6}$ \\
        & batch size & $100$ \\
        & return-normalizer & No \\
        \bottomrule
        \end{tabular}
    \end{subtable}\\
    \vspace{0.5cm}
 \begin{subtable}[t]{.5\textwidth}
     \centering
        \caption{Hopping MPO (muscle and torque)}
        \begin{tabular}{@{}lll@{}}
        \toprule
        &\textbf{Parameter} & \textbf{Value} \\
        \midrule
        & $\mathrm{lr}_{\mathrm{a}}$ & $3\times 10^{-4}$ \\
        & $\mathrm{lr}_{\mathrm{c}}$ & $3\times 10^{-4}$ \\
        & $\mathrm{lr}_{\mathrm{d}}$ & $10^{-2}$ \\
        & $\mathrm{clip}_{\mathrm{a}}$ & None \\
        & $\mathrm{clip}_{\mathrm{c}}$ & None \\
        & batch size & $256$ \\
        & return-normalizer & Yes \\
        \bottomrule
        \end{tabular}
    \end{subtable}\\
    \vspace{0.5cm}
       \begin{subtable}[t]{.5\textwidth}
        \centering
        \caption{Hopping perturbation (muscle)}
        \begin{tabular}{@{}lll@{}}
        \toprule
        &\textbf{Parameter} & \textbf{Value} \\
        \midrule
        & $\mathrm{lr}_{\mathrm{a}}$ & $9\times 10^{-4}$ \\
        & $\mathrm{lr}_{\mathrm{c}}$ & $3\times 10^{-3}$ \\
        & $\mathrm{lr}_{\mathrm{d}}$ & $10^{-2}$ \\
        & $\mathrm{clip}_{\mathrm{a}}$ & $10^{-5}$ \\
        & $\mathrm{clip}_{\mathrm{c}}$ & $3\times 10^{-7}$ \\
        & batch size & $256$ \\
        & return-normalizer & Yes \\
        \bottomrule
        \end{tabular}
    \end{subtable}%
    \begin{subtable}[t]{.5\textwidth}
        \centering
        \caption{Hopping perturbation (torque)}
        \begin{tabular}{@{}lll@{}}
        \toprule
        &\textbf{Parameter} & \textbf{Value} \\
        \midrule
        & $\mathrm{lr}_{\mathrm{a}}$ & $10^{-3}$ \\
        & $\mathrm{lr}_{\mathrm{c}}$ & $7\times 10^{-4}$ \\
        & $\mathrm{lr}_{\mathrm{d}}$ & $2\times 10^{-2}$ \\
        & $\mathrm{clip}_{\mathrm{a}}$ & $2\times 10^{-5}$ \\
        & $\mathrm{clip}_{\mathrm{c}}$ & $10^{-6}$ \\
        & batch size & $256$ \\
        & return-normalizer & Yes \\
        \bottomrule
        \end{tabular}
    \end{subtable}\\
    \vspace{0.5cm}
    \begin{subtable}[t]{.5\textwidth}
        \centering
        \caption{Common MPO settings}
        \begin{tabular}{@{}lll@{}}
        \toprule
        &\textbf{Parameter} & \textbf{Value} \\
        \midrule
        & buffer size  & $10^{6}$ \\
        & steps before batches & $5\times 10^{4}$ \\
        & steps between batches & $50$ \\
        & number of batches & $50$ \\
        & n-step return & $3$ \\
        & n parallel & $20$ \\
        & n sequential & $10$ \\
        \bottomrule
        \end{tabular}
    \end{subtable}%
\end{table*}
\section{Models}
We give detailed descriptions of the used models in this section. See \tab{tab:mujocomodels} for more information about the MuJoCo models. 
\label{supp:models}
\begin{table}
\caption{State information for all MuJoCo environments. The elements actuator lengths and velocities are directly derived from MuJoCo internal attributes \texttt{actuator\_length} and \texttt{actuator\_velocity} and keep the two morphologies as consistent as possible.}
\begin{tabular}{p{0.25\linewidth} p{0.68\linewidth}}
\textbf{model} & \textbf{observations}\\
 \toprule
 \armmujoco{} (muscle) & joint positions, joint velocities, muscle positions, muscle velocities, muscle forces, muscle activities, goal position, hand position\\
\armmujoco{} (torque) & joint positions, joint velocities, actuator positions, actuator velocities, actuator forces, goal position, hand position\\
\biped{} (muscle) & joint positions, joint velocities, muscle lengths, muscle velocities, muscle forces, muscle activities, head position, pelvis position, torso angle, scaled COM-velocity \\
\biped{} (torque) & joint positions, joint velocities, actuator lengths, actuator velocities, actuator forces, head position, pelvis position, torso angle, scaled COM-velocity\\
\bottomrule
\end{tabular}

\label{tab:mujocomodels}
\end{table}
\subsection{\arm}
\label{supp:arm26}
The \arm model consists of two segments connected with hinge joints moving against gravity.
The \armdemoa \citep{darus-2871_2022} is freely available using the multi-body software demoa \cite{darus-2550_2022}. In the muscle-actuated case, six muscles were included, modeled as Hill-Type muscles (\ref{sec:muscledemoa}). Here, two monoarticular muscles, each for the shoulder and elbow joint, and two biarticular muscles acting on both joints are included. The segments are modeled as rigid bodies, and the dynamics are solved using the Euler-Lagrange equation. In the torque-actuated case, each joint is driven by one torque actuator. For more details on the demoa model, we refer to the Technical Report \citep{darus-2871_2022}. The variant \armmujoco was derived from an implementation of Arm26 included in MuJoCo\,\cite{todorovMuJoCoPhysicsEngine2012a}, it was modified to yield a torque-variant similar to~\cite{schumacherdeprl2022}. We additionally created a muscle-variant consisting of 2 muscles per joint. The maximum torques for the torque actuators were matched to the highest achieved torques by the trained muscle policies for both versions independently.
\subsection{\biped}
\label{supp:biped}
We converted the geometrical model of an OpenSim bipedal human without arms~\cite{kidzinski_learning_2018-2} for use in MuJoCo. The model, consisting of 7 controllable joints (lower back, hip, knee, ankle) moves in a 2D-plane. Each joint is actuated by two antagonistic muscles or one idealized torque actuator. During execution, we only allow control signals for one leg, the actions for the other leg are kept identical to the first one. This incentivizes symmetric hopping motions, even though both legs can still move differently due to differing initial configurations or external forces. The maximum torques for the torque actuators were matched to the highest achieved torques by the trained muscle policies.  
\subsection{\allmin}
\label{supp:allmin}
For the squatting and high-jumping task, we used the \allmin  (allmin) model \citep{darus-2982_2022} which is freely available using the multi-body software demoa \cite{darus-2550_2022}. It consists of two legs and an upper body with a skeletal geometry similar to humans and moves in 3D. The ankle, knee and hip joints, as well as a lumbar and a cervical spine joint are controllable (8 controllable joints). The model also consists of two arms with their respective joints, however, these joints were not controlled in this study. In total, 14 joints are modeled with 20 degrees of freedom. Each controllable joint was either actuated by two muscles (\ref{sec:muscledemoa}) set up in an agonistic-antagonistic setup (muscle-actuated case) or by one idealized torque actuator (torque-actuated case). The maximum allowed torques were matched to the highest torques that occurred in the optimization in the muscle-actuated case to allow for a fair comparison.  Only monoarticular muscles (spanning one joint) were used. Furthermore, we reduced the number of control inputs $n_u$ for this study by using symmetrical control signals for the left and right legs. Additional to the torques generated by the actuators, also joint limitations are modeled as linear one-sided spring-damper elements. We refer to the Technical Report \citep{darus-2982_2022} for more details.
\section{Tasks}
\label{supp:tasks}
We chose movement objectives which represent both, robotic challenges and naturally observed movements of humans.

\paragraph{Smooth point-reaching (OC/MPC)} This task encourages smooth point-reaching. Therefore, the objective minimizes the L2-error between the desired angle endpoint and the desired joint angle velocity, as well as penalizing the angle jerk to ensure a smooth motion.
The objective for smooth point-reaching is given by:
\begin{equation}\label{eq:smoothpointreaching}
		\varepsilon = \frac{\omega_i}{{S_\mathrm{i}}}(\theta_\mathrm{i}-\thetades)^2+\frac{\omega_i}{{S_\mathrm{i}}}(\dot{\theta}_\mathrm{i}-\thetadotdes)^2+\dddot{\theta}^2,
\end{equation}
where $\theta_\mathrm{i}$ denotes the joint angle, $\dot{\theta}_\mathrm{i}$ the joint angle velocity and the last term $\dddot{\theta}$ penalizes the angle jerk to ensure a smooth motion. $\omega_i$ and $S_\mathrm{i}$ are weighting and scaling parameters, respectively. Their values (shoulder and elbow) are given in Table \ref{tab:scalingweightingparams}. The scaling parameters were chosen based on measured upper limits for human joint angular velocity \citep{jessop2016maximum} and human joint angle limits (Table 2 in \citep{darus-2982_2022}). The desired angle $\thetades$ is set to $90^{\circ}$ for both the shoulder (sh) and the elbow (elb) joint, as this requires a large motion. The movement duration in this task was set to $0.9$ s. 
\begin{table}
	\caption{Parameters for cost functions of OC/MPC tasks.}
	\centering
	\begin{subtable}[t]{.49\textwidth}
	\centering
	\caption{Scaling parameters}
	{\begin{tabular}{ll} \toprule
			parameter & value   \\  \midrule
			$S_{\theta,\textrm{sh}}$  & 2.45 [rad] \\
			$S_{\theta,\textrm{elb}}$ & 2.45 [rad] \\ 
			$S_{\theta,\textrm{hip}}$  & 1.92 [rad] \\
			$S_{\theta,\textrm{knee}}$ & 2.11 [rad] \\ 
			$S_{\theta,\textrm{ank}}$ & 1.05 [rad] \\ 
			$S_{\theta,\textrm{ls}}$ & 0.52 [rad] \\ 
			$S_{\theta,\textrm{cs}}$ & 1.05 [rad] \\ 
			$S_{\dot{\theta},\textrm{sh}}$  & 18.7 [rad/s]\\
			$S_{\dot{\theta},\textrm{elb}}$ & 27.9 [rad/s] \\
			$S_{\dot{\theta},\textrm{hip}}$  & 14.1 [rad/s]\\
			$S_{\dot{\theta},\textrm{knee}}$ & 28.4 [rad/s] \\ 
			$S_{\dot{\theta},\textrm{ank}}$ & 12.6 [rad/s] \\ 
			$S_{\dot{\theta},\textrm{ls}}$ & 5.2 [rad/s] \\ 
			$S_{\dot{\theta},\textrm{cs}}$ & 10.4 [rad/s] \\ \bottomrule
	\end{tabular}}
	\end{subtable}
	\begin{subtable}[t]{.49\textwidth}
	\centering
	\caption{Weighting parameters}
	{\begin{tabular}{ll} \toprule
			parameter & value   \\  \midrule
			$\omega_{\theta,\textrm{sh}}$ & 2  \\   
			$\omega_{\theta,\textrm{elb}}$ & 2  \\
			$\omega_{\theta,\textrm{hip}}$ & 2  \\   
			$\omega_{\theta,\textrm{knee}}$ & 2  \\
			$\omega_{\theta,\textrm{ank}}$ & 2  \\
			$\omega_{\theta,\textrm{ls}}$ & 2  \\
			$\omega_{\theta,\textrm{cs}}$ & 2  \\
			$\omega_{\dot{\theta},\textrm{sh}}$ & 1 \\ 
			$\omega_{\dot{\theta},\textrm{elb}}$ & 1 \\
			$\omega_{\dot{\theta},\textrm{hip}}$ & 1 \\ 
			$\omega_{\dot{\theta},\textrm{knee}}$ & 1 \\   
			$\omega_{\dot{\theta},\textrm{ank}}$ & 1 \\ 
			$\omega_{\dot{\theta},\textrm{ls}}$ & 1 \\
			$\omega_{\dot{\theta},\textrm{cs}}$ & 1  \\ \bottomrule
	\end{tabular}}
	\end{subtable}
	\label{tab:scalingweightingparams}
\end{table}
\paragraph{Precise point-reaching (RL)}
We employ a similar reward function to~\cite{haeufle2020muscles}:
\begin{equation}
    r = - \lambda_{1} (d - \log(d + \epsilon^{2})) - \frac{\lambda_{2}}{N}\sum a_{i}^{2} - 2,
\end{equation}
where $d$ is the Euclidean distance between end effector and target position, $\epsilon=10^{-4}$ prevents numerical instabilities, $\lambda_{1}=0.1$ and $\lambda_{2}=10^{-4}$. A smaller distance $d$ increases the overall reward, but in contrast to the usual Euclidean distance, the $\log$-term increases rewards for very small distances even further, incentivizing precision.
The episode does not terminate until a time limit of 1000 steps elapses. 

\paragraph{Fast point-reaching (RL)}
This task is identical to the previous one, but, in addition to the time limit, the episode also terminates if the distance between end effector and target position is below $5$ cm, which incentivizes reaching speed over precision.
\paragraph{Hitting a ball with a high velocity (OC/MPC)}
A ball with a mass of $250$ g is dropped in front of the arm model and the controller learns to hit the ball with a high velocity by optimizing the following objective:
\begin{equation}\label{eq:ballhitting}
		\varepsilon = -\max{\dot{z}_\text{ball}},
\end{equation}
where $\dot{z}_\text{ball}$ denotes the ball-velocity in z-direction (direction of gravity).
\paragraph{Squatting (OC/MPC)}
The objective for squatting is given by:
\begin{equation}\label{eq:squatting}
	\varepsilon = \frac{\omega_i}{{S_\mathrm{i}}}(\theta_\mathrm{i}-\thetades)^2+\frac{\omega_i}{{S_\mathrm{i}}}(\dot{\theta}_\mathrm{i}-\thetadotdes)^2,
\end{equation}
where $\theta_\mathrm{i}$ denotes the joint angle, $\dot{\theta}_\mathrm{i}$ the joint angle velocity. $\omega_i$ and $S_\mathrm{i}$ are weighting and scaling parameters, respectively. Their values are given in Table \ref{tab:scalingweightingparams}. The scaling parameters were chosen based on measured upper limits for human joint angular velocity \citep{jessop2016maximum} and human joint angle limits (Table 2 in \citep{darus-2982_2022}). The movement duration in this task was set to $0.9$ s. This squatting objective is taken from \cite{walter2021geometry}, where the desired hip $\thetadeship$, knee $\thetadesknee$ and ankle $\thetadesankle$ joint angle are defined to be:
\begin{align*}
	\thetadesankle &= -20^{\circ}, \\
	\thetadesknee &= \sin^{-1}(-\frac{L_s}{L_t} \cdot \sin(\thetadesankle))-\thetadesankle-\theta_{\text{an,}0}, \\
	\thetadeship &= -\thetadesknee-\thetadesankle. \\
\end{align*}
 \paragraph{High-Jumping (OC/MPC)}
 The objective for the high-jumping is taken from \citep{pandy1990optimal} and maximizes the position and velocity of the centre of mass of the human body model at the time of lift-off $t_l$. Additionally, we slightly expanded this objective to account for the three-dimensionality of our jumping model by penalizing deviations of the centre of mass in the $x$ and $y$-direction:
 \begin{equation}\label{eq:high-jumping}
	\varepsilon = z_\text{com}(t_l) + \frac{\dot{z}_\text{com}^2(t_l)}{2g} - |(x_\text{com}(t_l)-0)| - |(y_\text{com}(t_l)-0)|.
\end{equation}
Note, that $z_\text{com}$ denotes the centre of mass position (CoM) in z-direction (direction of gravity). The model is initialized to start from a squatting position in this task.

\paragraph{Hopping (RL)}
We developed a reward function that is able to induce hopping in different leg-driven systems and can be applied independently of the actuator morphology. We did not obtain good results with height-based rewards or the gym hopper~\cite{gym} reward function.
The reward for hopping is given by:
\begin{equation}
r = \exp(\mathrm{max}\{0, \hat{v}_{z}^{\mathrm{COM}}\}) - 1,
\end{equation}
where $v_{z}^{\mathrm{COM}}$ is the z-velocity of the center of mass. The transformation $\hat{v} = \mathrm{min}\{10, 100\,v\}$ adjusts the sensitivity of the reward function while also preventing numerical overflows of the exponential function. Crucially, large positive velocities are weighted much more strongly than small or negative velocities, driving the system to maximum height periodic hopping. The second term prevents positive rewards for velocities close to zero, as $\exp(0) = 1$.
We additionally use regularizing cost terms:
\begin{equation}
    r_{\mathrm{reg}} = r_{\mathrm{alive}} - \lambda_{1}\,r_{\mathrm{action}} - \lambda_{2}\, r_{\mathrm{joint}}, 
\end{equation}
where $\lambda_{1} = 10^{-4}$, $\lambda_{2}= 10^{-3}$, $r_{\mathrm{alive}}$ is $1$ if the episode does not terminate and $0$ otherwise, $r_{\mathrm{action}}=\sum a_{i}^{2}/N$ punishes large actions and $r_{\mathrm{joint}}$ punishes joint angles close to the limits of the system. Specifically:
\begin{equation}
    r_{\mathrm{joint}} = \begin{cases}
        -1, & \text{if } |q_{\mathrm{max},i} - q_{i}| < 0.1\\
        -1, & \text{if }|q_{\mathrm{min},i} - q_{i}| < 0.1\\
         0, & \text{otherwise}.
    \end{cases}
\end{equation}
Finally, we terminate the episode after the lapse of a time limit of 1000 iterations, or if different parts of the model are very close to the ground, as this indicates a fall. The termination conditions are:
\begin{align*}
    h_{\mathrm{skull}} &< 0.3 ~\mathrm{[m]}\\
    h_{\mathrm{pelvis}} &< 0.2~\mathrm{[m]}\\
    h_{\mathrm{tibia_l}} &< 0.3~\mathrm{[m]}\\
    h_{\mathrm{tibia_r}} &< 0.3~\mathrm{[m]}\\
    \theta_{\mathrm{torso}} &> 1.22~\mathrm{[rad]}\\
    \theta_{\mathrm{torso}} &< -0.88~\mathrm{[rad]}, 
\end{align*}
where $h$ is the height of the respective body part and $\theta_{\mathrm{torso}}$ marks the torso angle deviation from the upright position.
 \section{Additional experiments (RL)}
 \label{supp:additional}
\subsection{Maximum force variation}
\label{supp:forcevariation}
\begin{figure}
\centering
{\hspace{1.5cm}\small\textcolor{red}{\rule[2pt]{10pt}{1pt}} muscle \quad \textcolor{ourblue}{\rule[2pt]{10pt}{1pt}} torque}\\
\includegraphics[width=1.0\textwidth]{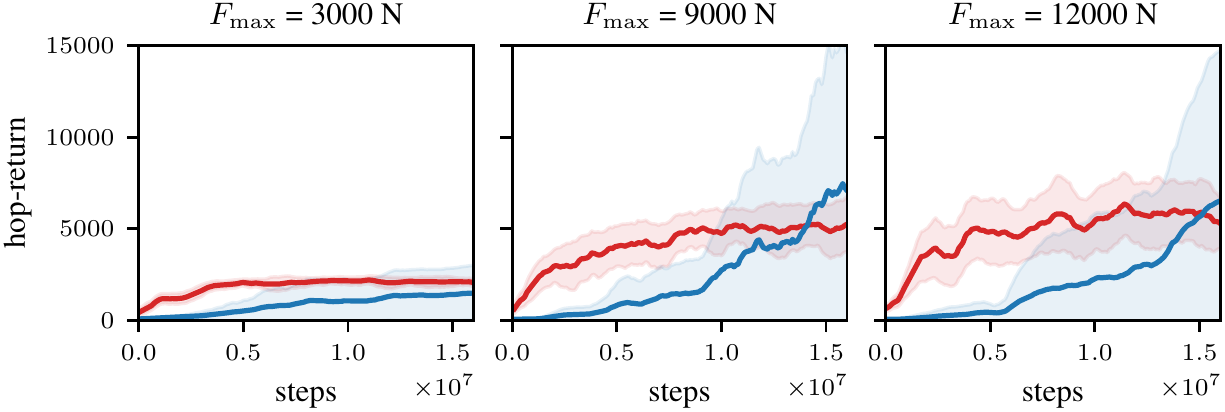}
	\caption{\textbf{Hopping performance for different actuator strengths.} Hyperparameters are optimized for hopping with a maximum muscle strengh of $F_{\mathrm{max}}=5000$ N as used in the previous hopping experiment. The maximum isometric muscle force is set to different values and the policies are trained for the task. Afterwards, the maximum used torques for the learned behaviors are recorded for each joint and set to identical values for the torque actuator. Muscle-actuators lead to more consistent performance and yield periodic hopping. Torque-actuators yield unstable policies that manage to jump very high once, but terminate the episode due to falls.}
\label{fig:forcevariation_hopping}
\end{figure}

Although the maximum force of the actuators is not freely adjustable in real systems, it is trivial to do so in simulation and has a strong influence on performance. We, therefore, used the biped parameters resulting from our hyperparameter optimization and recorded learning curves for the hopping task for both actuator morphologies for different maximum actuator forces. For each setting, we set the maximum isometric force for all muscle actuators to a certain value, trained the systems to convergence, and then recorded the torque values occurring at each controllable joint during execution of the hopping behavior. We then trained torque-actuator policies while setting $\tau_{\mathrm{max}}$ to the previously observed maximum values for \textit{each} individual joint. The results are shown in \fig{fig:forcevariation_hopping}. Even though singular torque-driven runs are able to outperform all muscle-driven runs at the end of training, this not only takes a considerable number of learning iterations, but also comes at the cost of strong learning instabilities. Looking at the learned behaviors, the torque-driven policies tend to jump very high, but violate the allowed torso-angles at the peak due to their unstable explorative policies. No periodic hopping could be observed. The muscle-driven policies, on the other hand, achieve periodic hopping, even though the apex hopping height is smaller.   
\subsection{Additional goals for point-reaching with perturbations}
\label{supp:moregoals}
\begin{figure}
\centering
\includegraphics[width=1.0\textwidth]{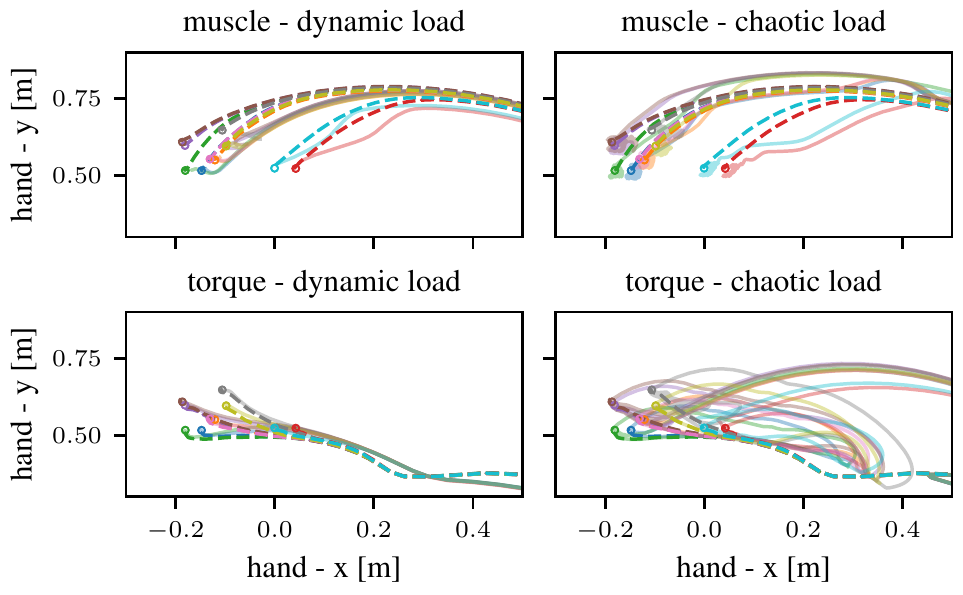}
	\caption{\textbf{Trajectories for dynamic ($\mathbf{1.5}$ kg weight) and chaotic (attached ball) load.} Left: The torque actuator handles the dynamic load case slightly better than the muscle actuator for all goals, especially compared to the goals in grey, brown and purple. Right: The muscle-actuator performs very well for all chaotic load goals, except for a small deviation from the end-point. The torque actuator exhibits strong instabilities. The respective goal positions are marked as circles, the unperturbed baseline for each goal is shown with a dashed line, the perturbed trajectories with slightly transparent solid lines.}
\label{fig:suppmoregoals}
\end{figure}

We show ten random arm goals for precise point-reaching with perturbations that were not present during training in \fig{fig:suppmoregoals}. 
\subsection{Additional hyperparameter variations}
\begin{figure}
\centering
\begin{subfigure}{1.0\textwidth}
\includegraphics[width=1.0\textwidth]{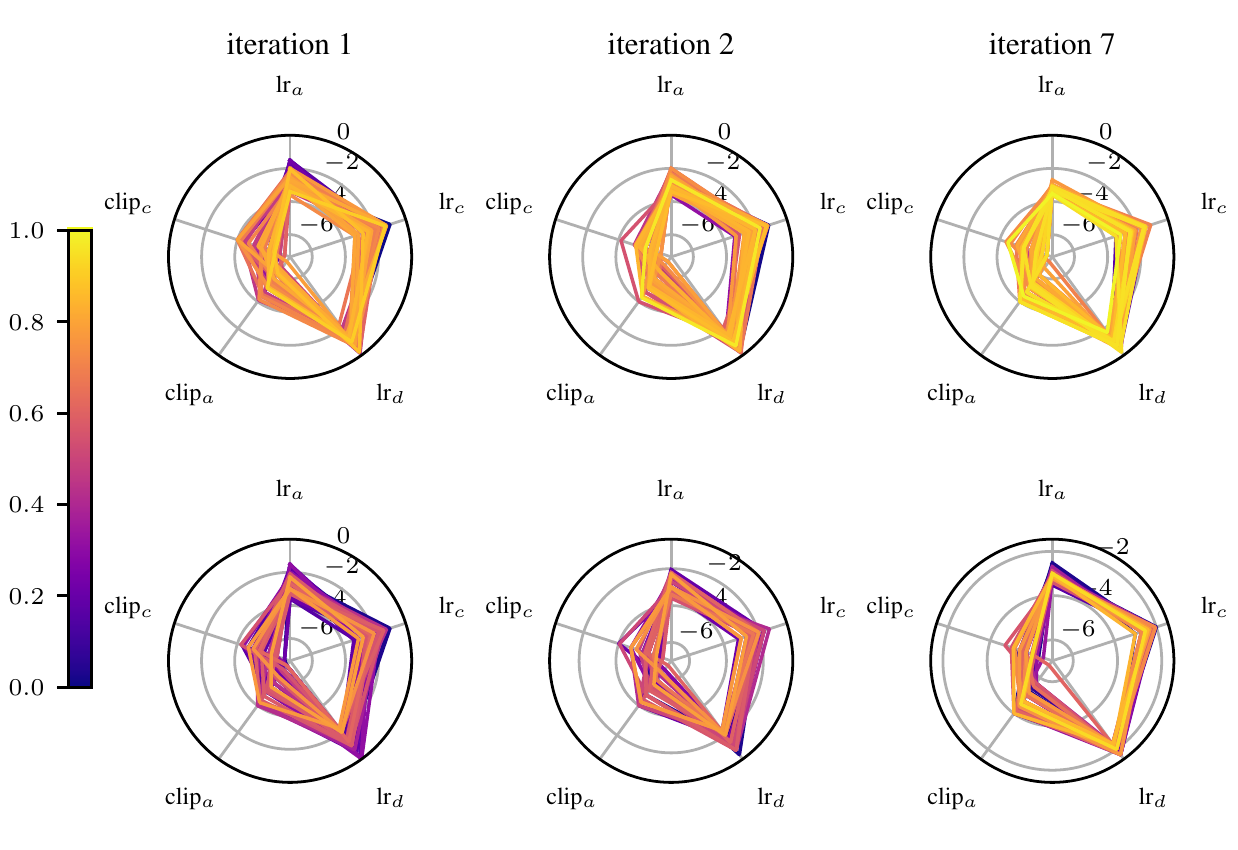}
\end{subfigure}\\
\begin{subfigure}{1.0\textwidth}
\hspace{0.23cm}
\includegraphics[width=1.0\textwidth]{figures/percentage_goodness_version2.pdf}
\end{subfigure}
	\caption{\textbf{Hyperparameter variation for precise point-reaching.} Hyperparameters are optimized following an iterative sampling scheme and individual runs train for $2\times 10^{6}$ iterations. Fifty sets of parameters are sampled randomly from pre-determined distributions, the final performance is evaluated and used to adapt the sampling distributions for the next iteration. We record 7 iterations which equals 350 runs in total. We optimize 5 parameters related to MPO. The angle of the radarplot marks the parameter, the radius marks the value (in $\log_{10}$-coordinates). Top: Radarplot of parameters for the muscle in precise point-reaching at iteration 1, 2 and 7. The color marks the achieved performance of the parameter sample relative to the best achieved performance over \textbf{all} sampled parameters. Middle: precise point-reaching torque. Bottom: Histogram of returns for all sets of parameters at iterations 1, 2 and 7.}
	\label{fig:supphyperpararlarm}
\end{figure}

\begin{figure}
\centering
\begin{subfigure}{1.0\textwidth}
\includegraphics[width=1.0\textwidth]{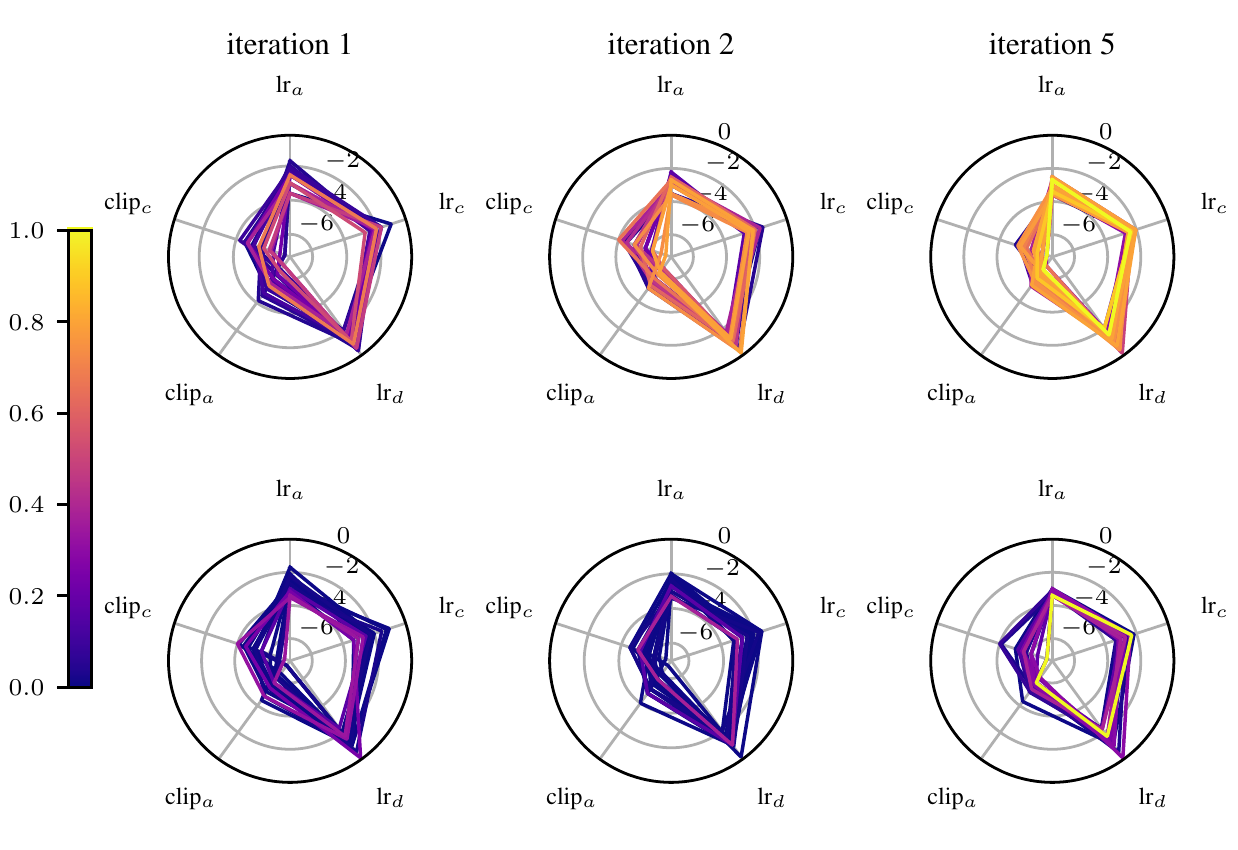}
\end{subfigure}\\
\begin{subfigure}{1.0\textwidth}
\hspace{0.24cm}
\includegraphics[width=1.0\textwidth]{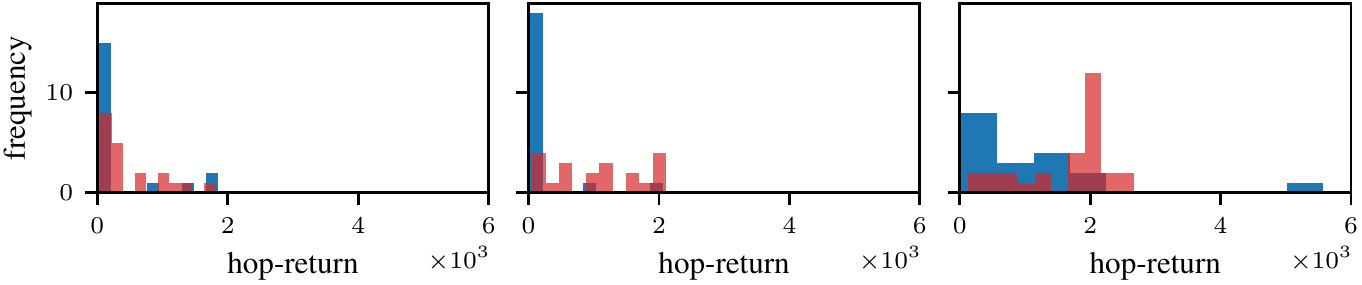}
\end{subfigure}
	\caption{\textbf{Hyperparameter variation for hopping.} Hyperparameters are optimized following an iterative sampling scheme and individual runs train for $5\times 10^{6}$ iterations. Twenty sets of parameters are sampled randomly from pre-determined distributions, the final performance is evaluated and used to adapt the sampling distributions for the next iteration. We record $5$ iterations which equals 100 runs in total. We optimize 5 parameters related to MPO. The angle of the radarplot marks the parameter, the radius marks the value (in $\log_{10}$-coordinates). Top: Radarplot of parameters for the muscle in the hopping task at iteration 1, 2 and 5. The color marks the achieved performance of the parameter sample relative to the best achieved performance over \textbf{all} sampled parameters. Middle: hopping torque. Bottom: Histogram of returns for all sets of parameters at iterations 1, 2 and 5.}
	\label{fig:supphyperpararlbiped}
\end{figure}

We show the relative performance of all runs of the hyperparameter searches in polar coordinates for precise point-reaching and hopping for both actuator morphologies (\fig{fig:supphyperpararlarm} and \fig{fig:supphyperpararlbiped}). The angles mark the specified hyperparameter (see \supp{supp:expdetails} for definitions), while the radius marks the chosen value in $\log_{10}$-coordinates. The top row marks performance with muscle-actuators, the middle row with torque-actuators and the bottom row shows histograms of returns for both morphologies at different iterations. For point-reaching, muscle morphology leads to a return distribution that is centered around the top-performing parameter sets, with almost no badly performing sets left at iteration 7. In contrast, for torque-morphology a large number of runs is still distributed at low return values. For hopping, a much harder task, muscle-morphology quickly leads to a large number of runs at the top-performance level, while some badly performing parameter sets remain even at iteration 5. For torque-morphology, a large peak can be observed for returns close to 0, as most sampled parameter sets do not achieve any kind of hopping. Only at iteration 7, a few singular well-performing runs appear, that strongly outperform even the best muscle-driven run. This was to be expected, as any muscle-actuator behavior can in principle be replicated by torque actuators, given that the policy is able to learn it. Muscle actuators, on the other hand, are restricted to trajectory-dependent output.
\subsection{Additional actuator models}
\label{sec:addactuatormodelsRL}
\begin{figure}
\centering

\includegraphics[width=1.0\textwidth]{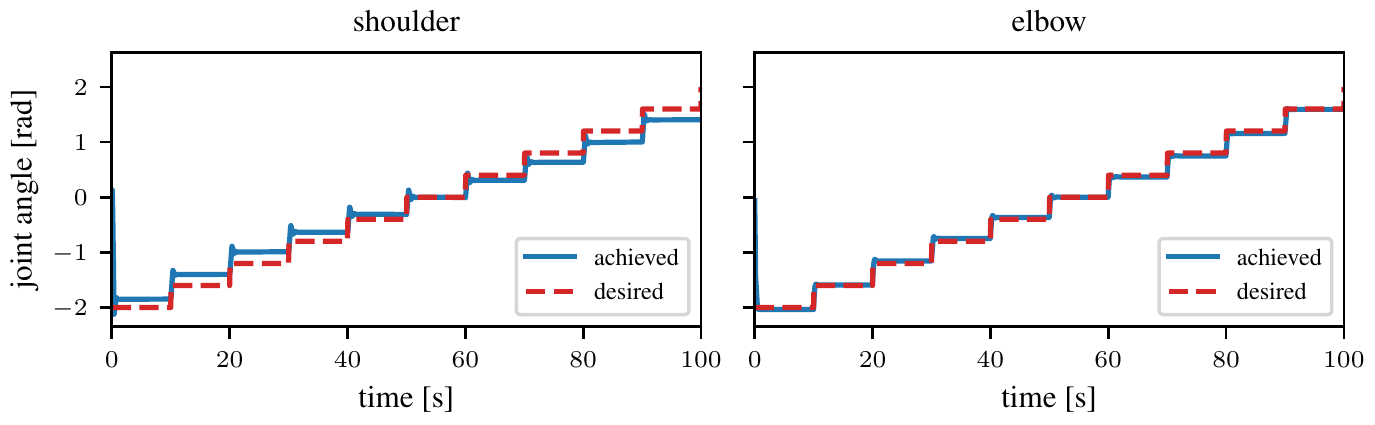}\\
\includegraphics[width=1.0\textwidth]{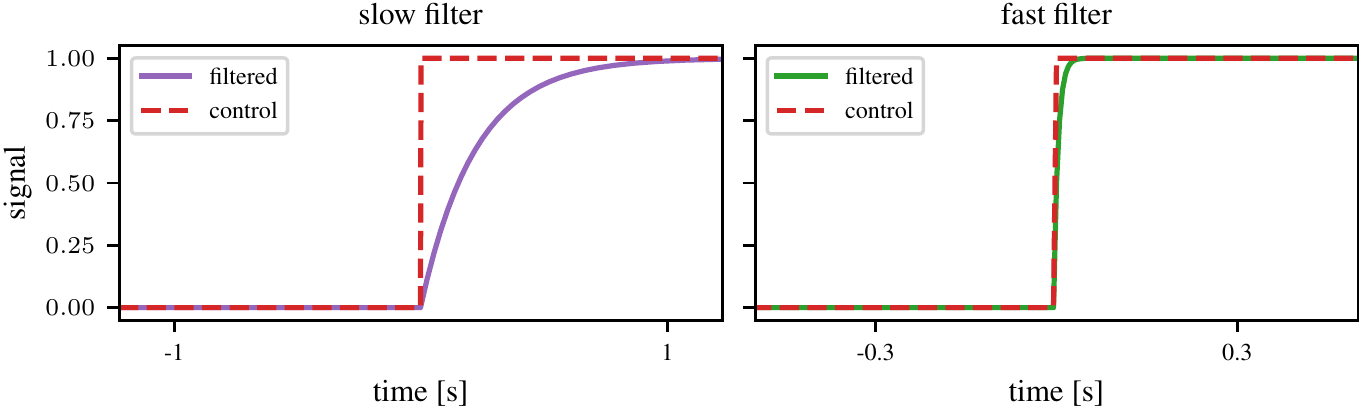}

	\caption{Top row: We tuned a PD-controller for \armmujoco that is then used as an intermediate control layer for an RL agent. We tuned the parameters by hand to achieve good joint angle control over the workspace, shown in the figure for both joint angles. The slight mismatch in the shoulder joint (left) is due to gravitational forces, which are not counteracted in the controller design. An RL agent easily learns to compensate for this shift. Bottom row: We show two low-pass filtered torque actuators for an exemplary step-signal. The fast-filter uses the same parameters as the activation dynamics in the MuJoCo muscle model.}
	\label{fig:supppd_controller}
\end{figure}

Similar to Peng et al.~\citep{peng2017learning}, we present more actuator models that are widely used in robotics. We consider the ideal torque actuator to be neutral in its properties---only executing exactly what it was told. In contrast, a PD-controller~\cite{pdcontrollerbook} embeds additional knowledge about position control elements and error propagation dynamics. For the RL experiments, we use an identical PD formulation to Peng et al.~\citep{peng2017learning}:
\begin{equation}
    u(t) = k_{p}\, (\hat{q}(t) - q(t)) + k_{d}\,(\hat{\dot{q}} - \dot{q}),
\end{equation}
with the joint angles $q$, the joint velocities $\dot{q}$, the desired position $\hat{q}$ and the desired velocity $\hat{\dot{q}}$. We also set $\hat{\dot{q}}=0$, similar to \citep{peng2017learning}. We tuned the PD-controller by hand to achieve good step-wise trajectory tracking, see \fig{fig:supppd_controller}. We also ensured that it remains stable for faster position changes. 

As a second additional actuator, we implemented a low-pass filtered torque actuator. The control signal is filtered according to the simplified muscle activation dynamics in the MuJoCo muscle model \eqn{eq:act_dyn_mujoco}, which effectively act as a low-pass filter:
\begin{equation}
\dot{a}(t) = \frac{1}{\Delta t} (u(t) - a(t)), 
\end{equation}
which gets approximated in practice as:
\begin{equation}
    a_{t+1} = a_{t} + \frac{\Delta t_{\mathrm{sim}}}{\Delta t} (u(t) - a(t)).
\end{equation}
The variable $a(t)$ denotes the effective action that is applied to the underlying torque actuator, $u(t)$ is the control signal, $\Delta t$ is the time scale of the low-pass filter and $\Delta t_{\mathrm{sim}}$ is the time step of the \textit{physics simulation}, which is not to be confused with the \textit{control time step}: $\Delta t_{\mathrm{control}} = 2\, \Delta t_{\mathrm{sim}}$ for the MuJoCo simulations. All actuator properties such as the muscle dynamics, the low-pass filter and the PD controller are updated with the physics simulation time step, while the RL policy computes new actions only with the control frequency.

\begin{figure}
\centering
\centering\large max torque = 30~Nm\\
\vspace{0.1cm}
{\hspace{1.2cm}\small\textcolor{red}{\rule[2pt]{10pt}{1pt}} muscle \quad \textcolor{ourblue}{\rule[2pt]{10pt}{1pt}} torque \quad \textcolor{pink}{\rule[2pt]{10pt}{1pt}} PD \quad\textcolor{purple}{\rule[2pt]{10pt}{1pt}} slow low-pass \quad \textcolor{green}{\rule[2pt]{10pt}{1pt}} fast low-pass}\\
\includegraphics[width=1.0\textwidth]{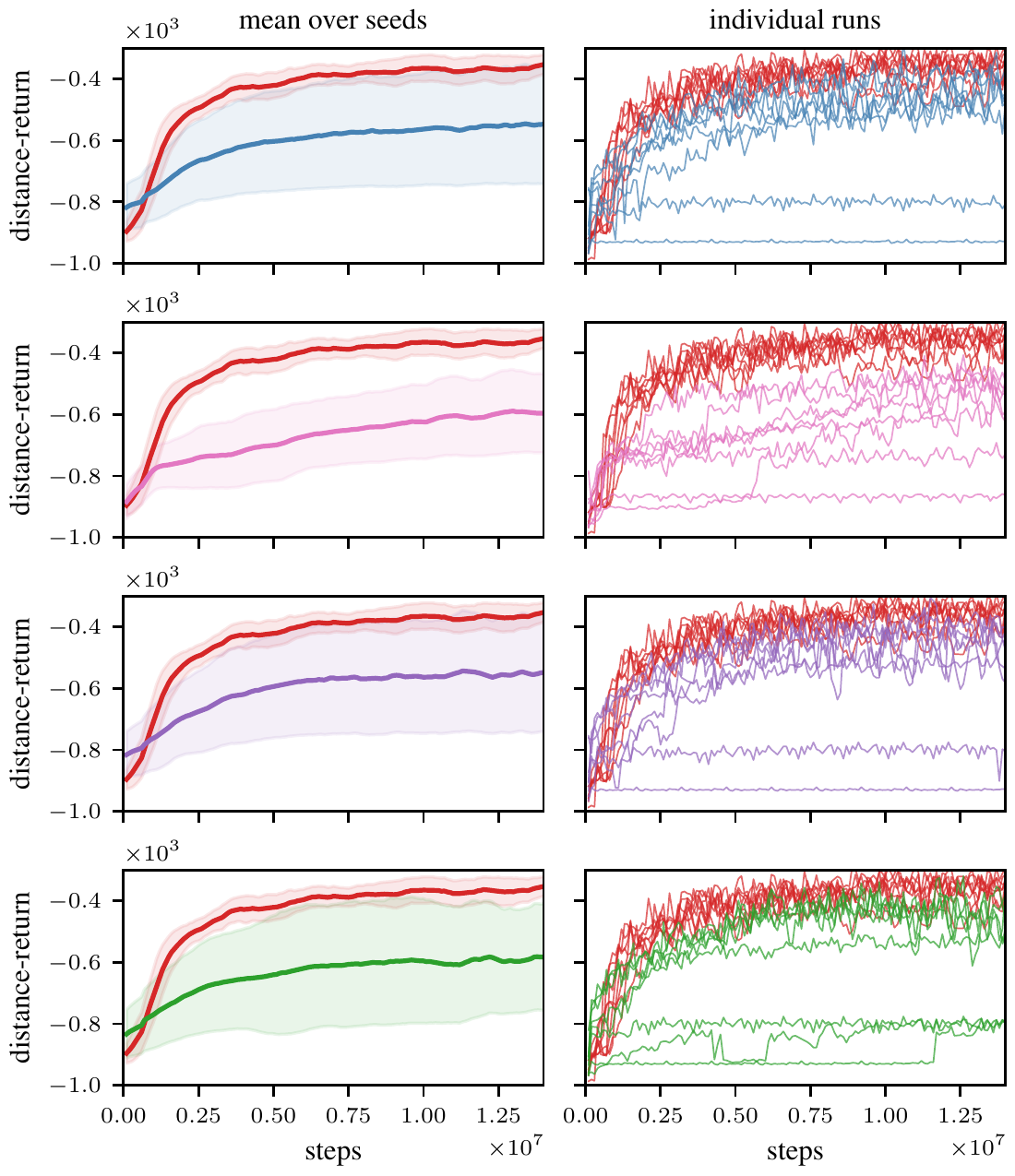}
	\caption{\textbf{The muscle actuator outperforms all other considered actuator designs.} We compare the learning curves for muscle, torque, PD and two low-pass filter actuators in the precise point-reaching task for \armmujoco. Averages across random seeds and standard deviation are shown in the left column, individual runs in the right column. The torque actuator and the low-pass filter variants perform quite well, but their variance across seeds is larger than for the muscle, even when outliers are not considered. The PD-controller seems to exhibit less variance than a pure torque-driven approach, but the overall performance is worse. We recorded 8 random seeds for each actuator.}
	\label{fig:learning_reaching_new_gear_30}
\end{figure}

We repeated the precise point-reaching task with \armmujoco with muscle actuation, torque actuation, PD actuation and two low-pass filter variants. The fast variant uses the time scale $\Delta t=0.01$, which is the same as used in the muscle model and reacts very fast to new control signals. The slow variant uses $\Delta t=1$ and produces a much stronger filtering effect. The results in \fig{fig:learning_reaching_new_gear_30} show that the muscle actuator outperforms all other variants. 

Individual runs are shown in the right column in order to obtain an accurate picture of the variance across seeds for all actuators. Even when not considering the badly performing outliers, torque actuation seems to present larger variance than muscle actuation. The PD-controller performs worse than pure torque control for this task, which validates results by \cite{learn2locomotepanne}: They found PD-controllers to perform worse than torque control when learning behaviors from scratch as opposed to tracking reference motions~\cite{peng2017learning}.

\begin{figure}
\centering
\centering\large max torque = 60~Nm\\
\vspace{0.1cm}
{\hspace{1.2cm}\small\textcolor{red}{\rule[2pt]{10pt}{1pt}} muscle \quad \textcolor{ourblue}{\rule[2pt]{10pt}{1pt}} torque \quad \textcolor{pink}{\rule[2pt]{10pt}{1pt}} PD \quad\textcolor{purple}{\rule[2pt]{10pt}{1pt}} slow low-pass \quad \textcolor{green}{\rule[2pt]{10pt}{1pt}} fast low-pass}\\
\includegraphics[width=1.0\textwidth]{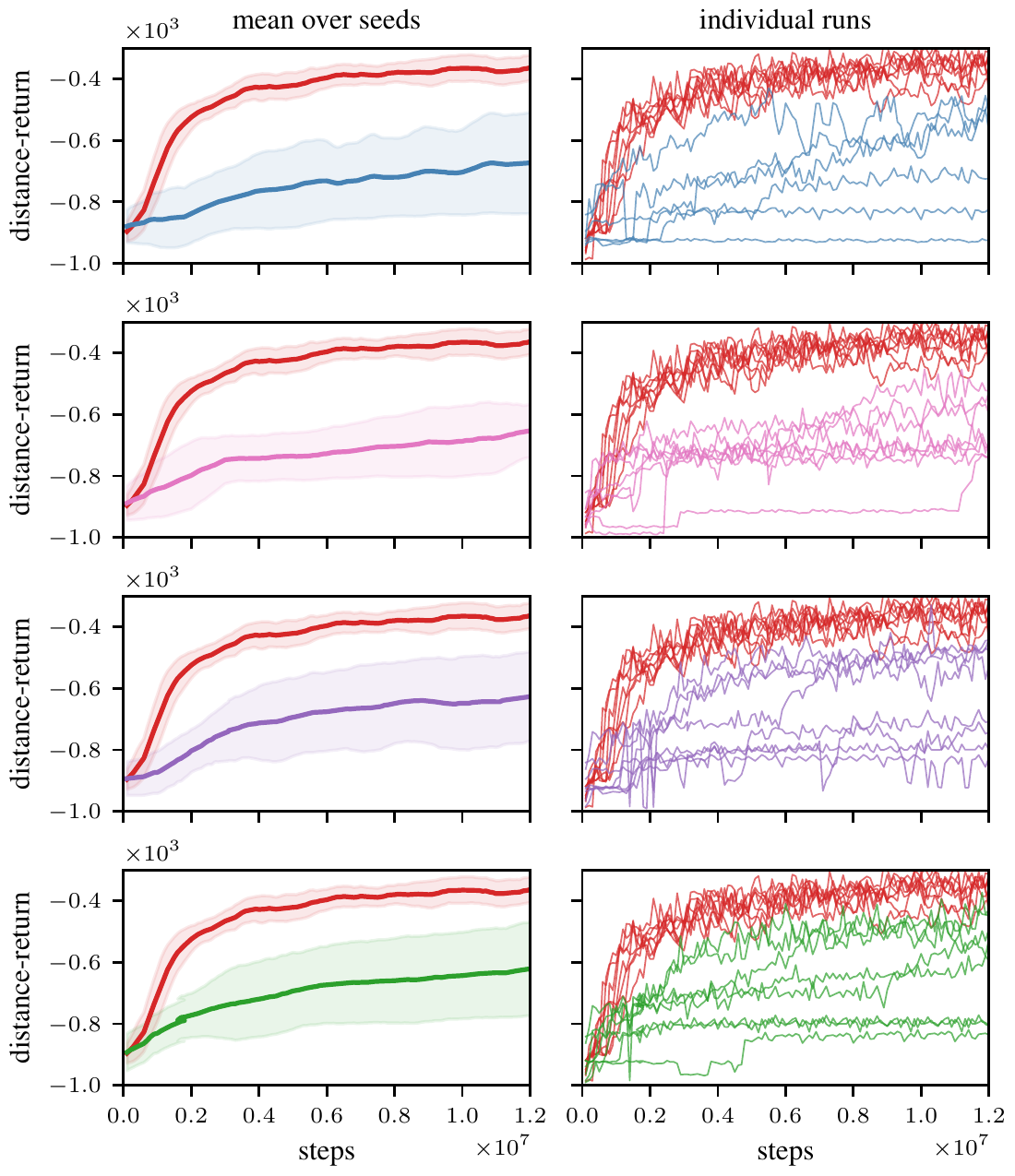}
	\caption{\textbf{Torque actuators perform worse when the maximum allowed force is increased.} We repeat the experiment in \fig{fig:learning_reaching_new_gear_60}, but allow the torque actuator to use a maximum torque of $\tau_{\mathrm{max}}=60$~Nm. This value is the maximum torque that the muscle actuator can output in perturbation experiments, even though it is not reached during point-reaching under normal conditions. While singular runs still perform well for the torque actuator variants and the PD controller achieves even less variance across seeds than before, the overall performance suffers when increasing the maximum force. We recorded 8 random seeds for each actuator.}
	\label{fig:learning_reaching_new_gear_60}
\end{figure}

We noticed that, while the muscle only uses a maximum of $\approx 30$~Nm during normal reaching, its properties allow it to intermittently use larger torques when perturbations are applied. We therefore conduct a second series of experiments where we adjust the maximum allowed torque for all torque actuators to the intermittent upper limit of the muscle, which is $\tau_{\mathrm{max}}=60$~Nm. New learning curves were recorded for \armmujoco point-reaching and are shown in \fig{fig:learning_reaching_new_gear_60}. Generally, the performance for the non-muscular actuators decreases with larger torque limits. Only the PD-controller seems to exhibit smaller variance than in the small torque limit case.

\subsection{Additional robustness experiments}
\begin{figure}
\centering
\centering\large max torque = 30~Nm\\
\includegraphics[width=1.0\textwidth]{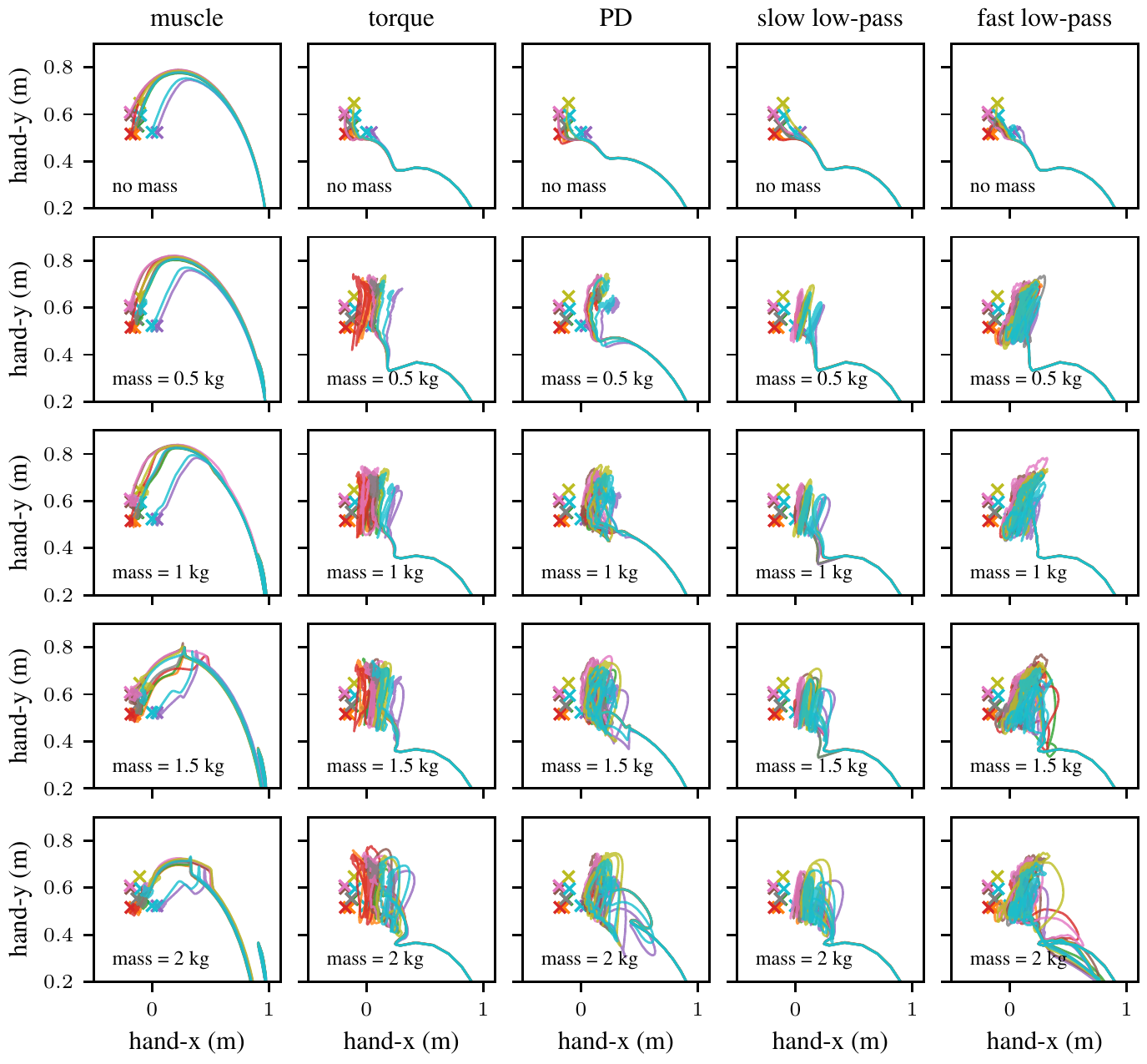}
	\caption{\textbf{The muscle actuator is more robust for all considered masses than the alternative.} We conducted perturbation experiments for all actuator models during which chaotic loads of differing masses were attached to the robot which were not present during training. The muscle actuator performs well up to $1.5$~kg, when deviations start to get bigger. It does not reliably reach the goal for $m=2$~kg. The torque actuator exhibits strong lateral oscillations for all masses and slight undershooting of the goal position. The PD-controller oscillates less for small masses, but undershoots the goals by a larger amount, as it was not tuned for this scenario. The low-pass filtered actuators perform similar to the pure torque case. For each experiment we used the best performing policy of each learning curve in \fig{fig:learning_reaching_new_gear_30} at the end of training. Ten goals were randomly chosen and used for all experiments.}
	\label{fig:suppperturbation_gear30}
\end{figure}
\begin{figure}
\centering
\centering\large max torque = 60~Nm\\
\includegraphics[width=1.0\textwidth]{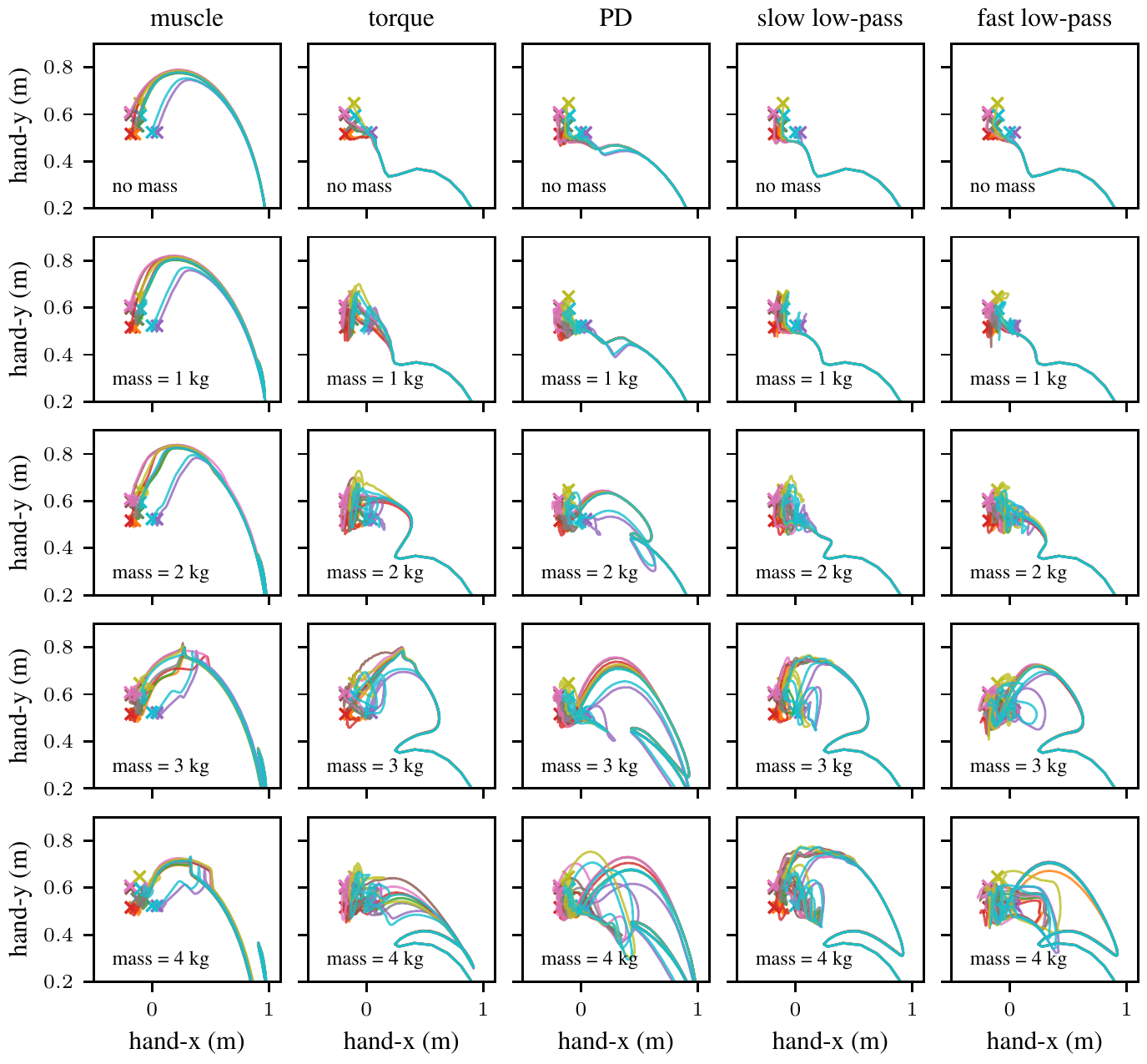}
	\caption{\textbf{Torque control is more robust with larger torque limits, but is still outperformed by muscles.} We repeat the experiment in \fig{fig:suppperturbation_gear30} with a larger maximum torque limit of $60$~Nm. All torque-variants seem to perform better than in the low torque limit case. For $m=1$~kg, the PD-controller and the low-pass filter versions slightly outperform the pure torque actuator. Nevertheless, the muscle reacts more robustly for all considered masses. For each experiment we used the best performing policy of each learning curve in \fig{fig:learning_reaching_new_gear_60} at the end of training. The same ten goals as in \fig{fig:suppperturbation_gear30} were used.}
	\label{fig:suppperturbation_gear60}
\end{figure}
In this section, we present evaluation of the robustness of the learned policies with a wide variety of masses and additional actuators. The results are reported for two different maximum torque limits for the non-muscle-based actuators, following the reasoning of \sec{sec:addactuatormodelsRL}.

The variations are investigated for policies trained for point-reaching with \armmujoco. All weights are added as a chaotic load that is attached with a rope. The results can be seen in \fig{fig:suppperturbation_gear30} and \fig{fig:suppperturbation_gear60} for $\tau_{\mathrm{max}}=30$~Nm and $\tau_{\mathrm{max}}=60$~Nm respectively. We use masses varying from 1 to 4 kg in the high force case, while they are halved in the other case. Even though the muscle actuator is the most stable across all variations, the pure torque actuator variant performs quite well when large forces are allowed. However, large torque limits also diminish the learning performance, as seen previously in \fig{fig:learning_reaching_new_gear_60}. The results suggest a trade-off between learning speed and robustness for the torque controller, while the muscle actuator is able to leverage low forces during learning and automatically reacts to perturbations with stronger forces. The PD-controller only outperforms raw torque control for the large torque limit $\tau_{\mathrm{max}}=60$~Nm and a comparatively small perturbation mass of $1$ kg, see \fig{fig:suppperturbation_gear60} (second row, middle).

\subsection{MuJoCo simulation time step ablation}
\begin{figure}
\centering
\centering\large max torque = 30~Nm\\
\includegraphics[width=1.0\textwidth]{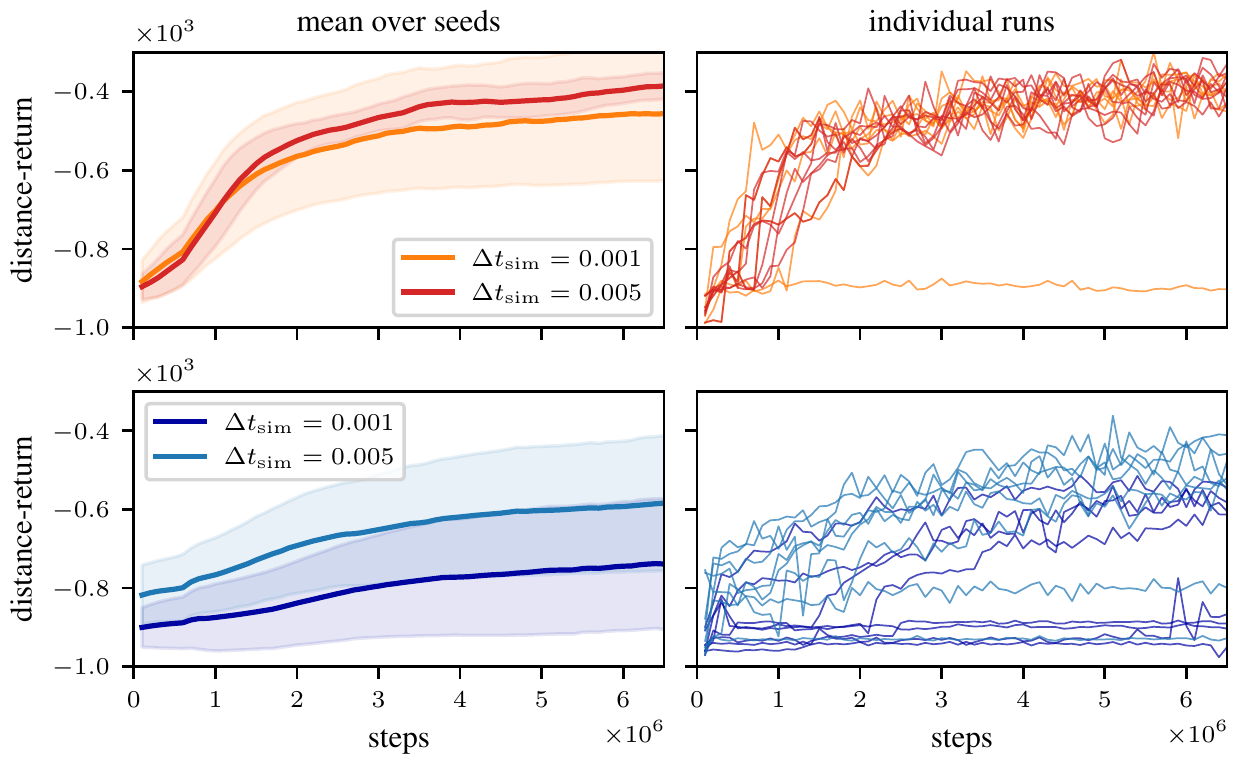}
	\caption{We present physical simulation time step ablations for \armmujoco and precise point-reaching. While the $\Delta t_{\mathrm{sim}}$ was varied, the frameskip parameter was adjusted to achieve the same control time step $\Delta t_{\mathrm{control}}=0.01$ in all cases. Top: With the exception of a single unlucky run, the simulation time step does not seem to affect the performance of the muscle actuator in this task. Bottom: Surprisingly, the torque actuator performs much worse with a \textit{smaller} simulation time step. As simulation accuracy increases with a smaller time step, we do not suspect this to be the result of numerical instability. A setting of $\Delta t_{\mathrm{sim}}=0.005$ was used for the all other MuJoCo experiments. We recorded 8 random seeds for each actuator.}
	\label{fig:timestep_ablation_mujoco}
\end{figure}

To assess the influence of simulation accuracy on the obtained results, we record additional muscle and torque actuator learning curves with a much smaller simulation time step of $\Delta t_{\mathrm{sim}}=0.001$ instead of $\Delta t_{\mathrm{sim}}=0.005$. We additionally increase the frameskip of the simulation to achieve an equal control time step of $\Delta t_{\mathrm{control}}=0.01$ in both cases. The results are shown in \fig{fig:timestep_ablation_mujoco}. With the exception of one unlucky run, the muscle actuator performance seems to be unchanged under the more accurate simulation setting. The torque actuator, in contrast, seems to perform worse. As simulation accuracy increases with a smaller time step, we do not suspect this to be the result of numerical instability. As this only reinforces prior results, we conclude that the improved muscle actuator data-efficiency is not a result of numerical instability.

\section{Additional experiments (OC/MPC)}
\subsection{Nonlinearity in muscle model}
\begin{figure}
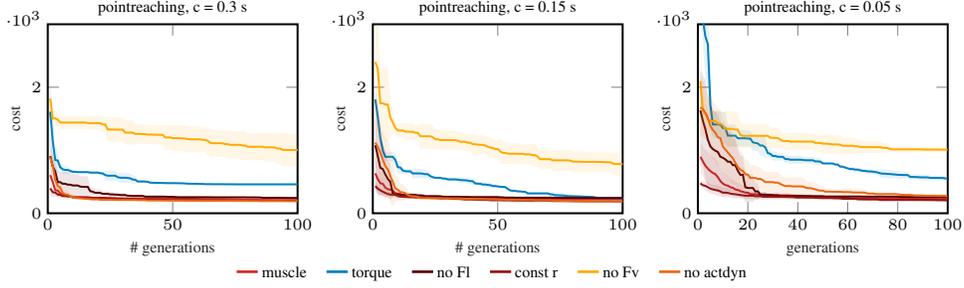

	\tikzsetnextfilename{compare_dataefficiency_musclemodifs}
	\centering
	\tiny
	\pgfplotsset{
		compat=1.6,
		try min ticks = 3,
		legend image code/.code={
			\draw[mark repeat=2,mark phase=2]
			plot coordinates {
				(0cm,0cm)
				(0.15cm,0cm)        %
				(0.3cm,0cm)         %
			};%
		}
	}
    \definecolor{mycolor1}{rgb}{0.00000,0.50000,0.75000}%
    \definecolor{mycolor2}{rgb}{0.55000,0.66250,0.94375}%
    \definecolor{mycolor3}{rgb}{0.77500,0.55000,0.55000}%
	\definecolor{mycolor1}{rgb}{0.00000,0.50000,0.75000}%
\definecolor{mycolor2}{rgb}{0.33333,0.00000,0.00000}%
\definecolor{mycolor3}{rgb}{1.00000,0.66667,0.00000}%
\begin{tikzpicture}
\def\Factor{-3}
\setlength{\figH}{0.18\textwidth}
\setlength{\figW}{0.25\textwidth}
\input{new_figures/dataefficiency_cost_pointreaching_modifs0_3}
\input{new_figures/dataefficiency_cost_pointreaching_modifs0_15}
\input{new_figures/dataefficiency_cost_pointreaching_modifs0_05}
\draw (5.9, -0.8) node {\tiny\textcolor{red}{\rule[1.5pt]{8pt}{1pt}} muscle \quad \textcolor{mycolor1}{\rule[1.5pt]{8pt}{1pt}} torque \quad \textcolor{mycolor2}{\rule[1.5pt]{8pt}{1pt}} no Fl \quad \textcolor{red!50!mycolor2}{\rule[1.5pt]{8pt}{1pt}} const r \quad
\textcolor{mycolor3}{\rule[1.5pt]{8pt}{1pt}} no Fv \quad
\textcolor{red!50!mycolor3}{\rule[1.5pt]{8pt}{1pt}} no actdyn}; %
\end{tikzpicture}%
	\caption{\textbf{Cost value for point-reaching while switching off different muscle properties.} Plotting the mean and standard deviation (shaded area) for 5 repeated runs for the two main actuator morphologies (muscle in red, torque in blue). Additionally, different morphologies are tested where muscle properties are switched off separately: We switched off the force-length relation (no Fl), set moment arms to be constant (const r), switched off the force-velocity (no Fv) and excluded the activation dynamics (no actdyn). }
	\label{fig:data_efficiency_nonlinearmuscleproperties}
\end{figure}

In our study, we conclude that the nonlinear muscle properties can be beneficial for learning in terms of data-efficiency and robustness. To show-case the influence of individual properties, we performed additional smooth point-reaching and squatting experiments. The four major properties that differ between the torque actuator morphology and the muscle actuator morphology are the nonlinear activation dynamics, the nonlinear force-length, the nonlinear force-velocity relation and the nonlinear lever arms (see also Fig. \ref{fig:muscle_flva_plot} in the main paper). We switched each of these properties separately off to test which nonlinear muscle property contributes the most to the beneficial behavior. The results can be seen in Fig.~\ref{fig:data_efficiency_nonlinearmuscleproperties}. As shown in this figure, switching off the nonlinear force-velocity relation (no Fv) has the strongest impact and leads to results that are even worse than the torque actuator optimization. Additionally, the nonlinear activation dynamics (no actdyn) has some influence on the performance of the data-efficiency results. With these results, we would like to give a first indication that indeed the non-linearity of different muscle properties are beneficial for the data-efficiency in learning anthropomorphic tasks.

\color{black}

\subsection{Proportional-derivative torque control for learning}
\label{supp:pd}
\begin{figure}
	\tikzsetnextfilename{compare_dataefficiency_muscletorquepd}
	\centering
	\tiny
	\pgfplotsset{
		compat=1.6,
		try min ticks = 3,
		legend image code/.code={
			\draw[mark repeat=2,mark phase=2]
			plot coordinates {
				(0cm,0cm)
				(0.15cm,0cm)        %
				(0.3cm,0cm)         %
			};%
		}
	}
	\definecolor{mycolor1}{rgb}{0.00000,0.50000,0.75000}%
\begin{tikzpicture}
\def\Factor{-3}
\setlength{\figH}{0.24\textwidth}
\setlength{\figW}{0.42\textwidth}
\input{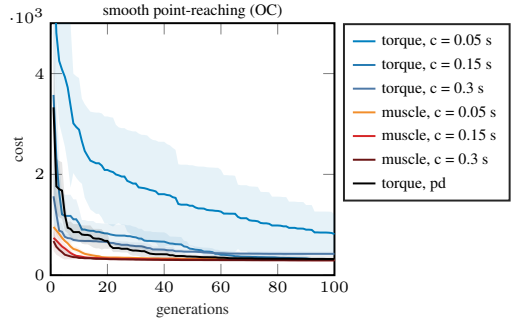}
\end{tikzpicture}%
	\caption{\textbf{Cost value for smooth point-reaching with additional baseline using PD control for torque actuator.} Plotting the mean and standard deviation (shaded area) for 5 repeated runs for the three main actuator morphologies (muscle in red, torque in blue, torque with PD controller in black).}
	\label{fig:data_efficiency_muscletorqpd}
\end{figure}

In the results presented in the main paper, we mainly compared the muscle actuator morphology to an idealized torque actuator without embedding any additional knowledge, e.g. position control which is typically used with a PD controller. Nevertheless, we consider the comparison with the PD control action space as a valuable baseline comparison. Therefore, we performed additional experiments for the smooth point-reaching task, where we added this additional baseline using PD control on top of the torque actuator morphology. 
Similar to the RL experiments (\ref{sec:addactuatormodelsRL}), we use an identical PD formulation to Peng et al.~\citep{peng2017learning}: 
\begin{equation}
    u(t) = k_{p}\, (\hat{q}(t) - q(t)) + k_{d}\,(\hat{\dot{q}} - \dot{q}),
\end{equation}
with the joint angles $q$, the joint velocities $\dot{q}$, the desired position $\hat{q}$ and the desired velocity $\hat{\dot{q}}$. We also set $\hat{\dot{q}}=0$, similar to \citep{peng2017learning} and our original cost function for smooth point-reaching (Eq. \ref{eq:smoothpointreaching}). In contrast to the RL experiments where we directly learn the desired angles for the control signal $u(t)$ for PD controller, here, with OC, we instead learn the $k_p$ and $k_d$ parameters. We allow for changes in these parameters every $c = 0.3$ s, whereas the control signal was updated continuously. 
Fig.~\ref{fig:data_efficiency_muscletorqpd} shows that the data-efficiency is slightly improved using a PD controller for the torque-actuated case in the smooth point-reaching task but it does not reach the performance of the muscle actuator.

\color{black}

\subsection{Additional robustness experiments}
\label{sec:mpc_more_weights}
\begin{figure}
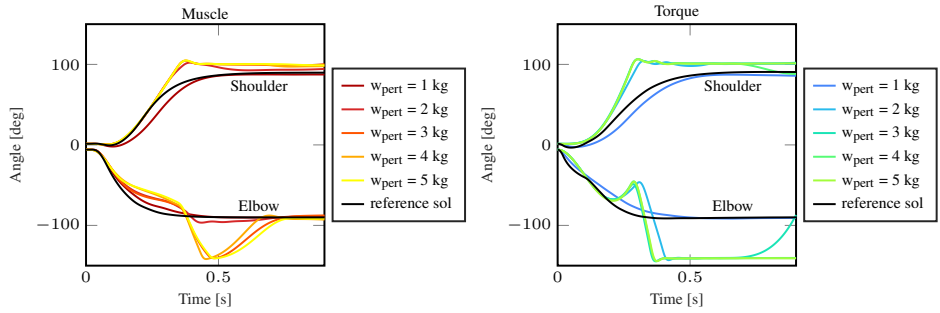

	\tikzsetnextfilename{compare_robustness_muscletorq_varyweightpointreach}
	\centering
	\tiny
	\pgfplotsset{
		compat=1.6,
		try min ticks = 3,
		legend image code/.code={
			\draw[mark repeat=2,mark phase=2]
			plot coordinates {
				(0cm,0cm)
				(0.15cm,0cm)        %
				(0.3cm,0cm)         %
			};%
		}
	}
	\definecolor{mycolor1}{rgb}{0.66667,0.00000,0.00000}%
\definecolor{mycolor2}{rgb}{1.00000,0.33333,0.00000}%
\definecolor{mycolor3}{rgb}{1.00000,0.66667,0.00000}%
\definecolor{mycolor4}{rgb}{1.00000,1.00000,0.00000}%
\begin{tikzpicture}
\setlength{\figH}{0.23\textwidth}
\setlength{\figW}{0.32\textwidth}
\input{new_figures/robustness_traj_varyweight_musc}
\definecolor{mycolor1}{rgb}{0.27327,0.52554,0.98180}%
\definecolor{mycolor2}{rgb}{0.16015,0.73318,0.92519}%
\definecolor{mycolor3}{rgb}{0.10342,0.89600,0.71500}%
\definecolor{mycolor4}{rgb}{0.33899,0.98295,0.45994}%
\definecolor{mycolor5}{rgb}{0.63843,0.99097,0.23646}%
\input{new_figures/robustness_traj_varyweight_torq}
\draw (2.3, 2.4) node {\tiny\textcolor{black} Shoulder};
\draw (2.3, 0.8) node {\tiny\textcolor{black} Elbow};
\draw (8.6, 2.4) node {\tiny\textcolor{black} Shoulder};
\draw (8.6, 0.8) node {\tiny\textcolor{black} Elbow};
\end{tikzpicture}%
	\caption{\textbf{Muscle morphology is more robust towards unknown weight perturbations.}  Plotting the angle trajectories of the shoulder and elbow angle over time for the two actuator morphologies (left: muscle, right: torque) while varying the unknown weight (between $1$ and $5$ kg in $1$ kg steps).  }
	\label{fig:robustness_varyweightpointreach}
\end{figure}

In this section, we present additional robustness experiments for perturbing the arm model in the point-reaching task while adding unknown weights to the lower arm. In contrast to the main paper, we do not only show the perturbation using $1$ kg, but varied the unknown weight up until $5$ kg in $1$ kg steps. The resulting angle trajectories are shown in Fig. \ref{fig:robustness_varyweightpointreach}. We see that both actuators are able to counteract unknown perturbation weights with $1$ kg. For larger weights, the perturbations result in overshoots in the elbow joint angle which can be corrected in the muscle-actuated case, whereas the torque actuator struggles to counteract these perturbations. Summed up, the muscle morphology is more robust towards perturbations for a wide range of different unknown weights. 
\color{black}

\end{document}